\documentclass[pmlr]{jmlr}

\RequirePackage{graphicx}
 \usepackage{booktabs}

\usepackage{tikz}

\usepackage{colortbl}
\usepackage{longtable}
\usepackage{multirow}
\usepackage{algorithm}
\usepackage{algpseudocode}
\makeatletter
\def\set@curr@file#1{\def\@curr@file{#1}} 
\makeatother
\usepackage[load-configurations=version-1]{siunitx} 

\theorembodyfont{\upshape}
\theoremheaderfont{\scshape}
\theorempostheader{:}
\theoremsep{\newline}

\definecolor{mygreen}{HTML}{34FF34}
\jmlrvolume{[VOLUME \# TBD]}
\jmlryear{2024}
\jmlrworkshop{Machine Learning for Healthcare}

\title[Medical imaging Enhanced with Dynamic Self-Adaptive Semantic Segmentation]{Exploring Intrinsic Properties of Medical Images for
Self-Supervised Binary Semantic Segmentation}

\author{\Name{Pranav singh}
       \Email{ps4364@nyu.edu}\\ 
       \addr Center for Data Science\\
       New York University\\
       New York, NY 10011, USA
       \AND
       \Name{Jacopo Cirrone}
       \Email{cirrone@courant.nyu.edu}\\ 
       \addr Center for Data Science 
        and Colton Center for Autoimmunity\\
        New York University\\
        New York, NY 10011, USA}

\begin{document}

\maketitle

\begin{abstract}
  Recent advancements in self-supervised learning have unlocked the potential to harness unlabeled data for auxiliary tasks, facilitating the learning of beneficial priors. This has been particularly advantageous in fields like medical image analysis, where labeled data are scarce. Although effective for classification tasks, these approaches have not been studied for complex applications, such as medical image segmentation. In this paper, we benchmark existing state-of-the-art self-supervised techniques for medical image segmentation on four challenging datasets of different modalities. Furthermore, we introduce \textbf{M}edical imaging \textbf{E}nhanced with \textbf{D}ynamic \textbf{S}elf-\textbf{A}daptive \textbf{S}emantic \textbf{S}egmentation (MedSASS), a dedicated end-to-end self-supervised framework tailored for medical image segmentation. We evaluate MedSASS against existing state-of-the-art methods across four diverse medical datasets, showcasing its superiority. With encoder only training, MedSASS outperforms existing CNN-based self-supervised methods by 3.83\% and matches the performance of ViT-based methods. Furthermore, when MedSASS is trained end-to-end, it demonstrates significant improvements of 14.4\% for CNNs and 6\% for ViT-based architectures compared to existing state-of-the-art self-supervised strategies.
\end{abstract}

\section{Introduction}

The field of medical image analysis represents a significant application of deep learning, typically involving tasks such as classification and segmentation. Classification entails categorizing each image into one or more classes. In contrast, segmentation, a more intricate task, requires categorizing each pixel in an image. Among these, semantic segmentation, which involves differentiating the image into foreground and background, is particularly crucial in medical imaging. This process is essential for isolating objects of interest (e.g., lesions in skin lesion imaging or cells in histopathology slides) from irrelevant areas, aiding clinicians in making more informed decisions.

Artificial intelligence has markedly enhanced the effectiveness of medical image analysis tools. A key principle in this domain is that AI-driven methods enhance performance proportionally to the volume of labeled data used in training. However, acquiring medical imaging data can be costly, time-consuming, and subject to varying levels of bureaucratic approval, posing challenges to the release of large-scale, well-labeled datasets. Recently, improvements in self-supervised learning methods have made it possible to use unlabeled data to learn priors, which can then be used to improve the performance of downstream tasks. During pre-training self-supervised techniques, perform an auxiliary task on unlabeled data to learn priors. Existing state-of-the-art self-supervised approaches only focus on training an encoder. Then this encoder can be fine-tuned for various downstream tasks such as segmentation, classification, object detection, image retrieval and so on. Self-supervised learning has shown encouraging results for classification of medical images \cite{azizi2021big,pmlr-v219-singh23a} on a variety of medical imaging datasets. But the performance of existing state-of-the-art self-supervised approaches haven't been studied on different modalities of medical imaging datasets.
With this work we start by benchmarking existing state-of-the-art self-supervised approaches for image segmentation on different modalities of medical imaging - Histopathology, Dermatology and Chest X-Ray images. In Section \ref{desc-of-MedSASS} we introduce MedSASS (\textbf{M}edical imaging \textbf{E}nhanced with \textbf{D}ynamic \textbf{S}elf-\textbf{A}daptive \textbf{S}emantic \textbf{S}egmentation (MedSASS)) - a novel self-supervised approach for medical image segmentation. Existing self-supervised techniques only train an encoder during pre-training. But effective image segmentation requires an effective encoder as well as a decoder. With MedSASS, we have the flexibility of either encoder only pre-training or end-to-end (encoder and decoder) training. For fair comparison, we start with encoder only pre-training in Section \ref{encoder-only}, for which MedSASS outperforms existing CNN-based self-supervised methods by 3.83\% and matches the performance of ViT-based methods. With end-to-end pre-training it demonstrates significant improvements of 14.4\% for CNNs and 6\% for ViT-based architectures compared to existing state-of-the-art self-supervised strategies. We present these results in Section \ref{enc-decoder-results}. 


\begin{figure}[ht]
\begin{center}
\centerline{\includegraphics[width=1\linewidth]{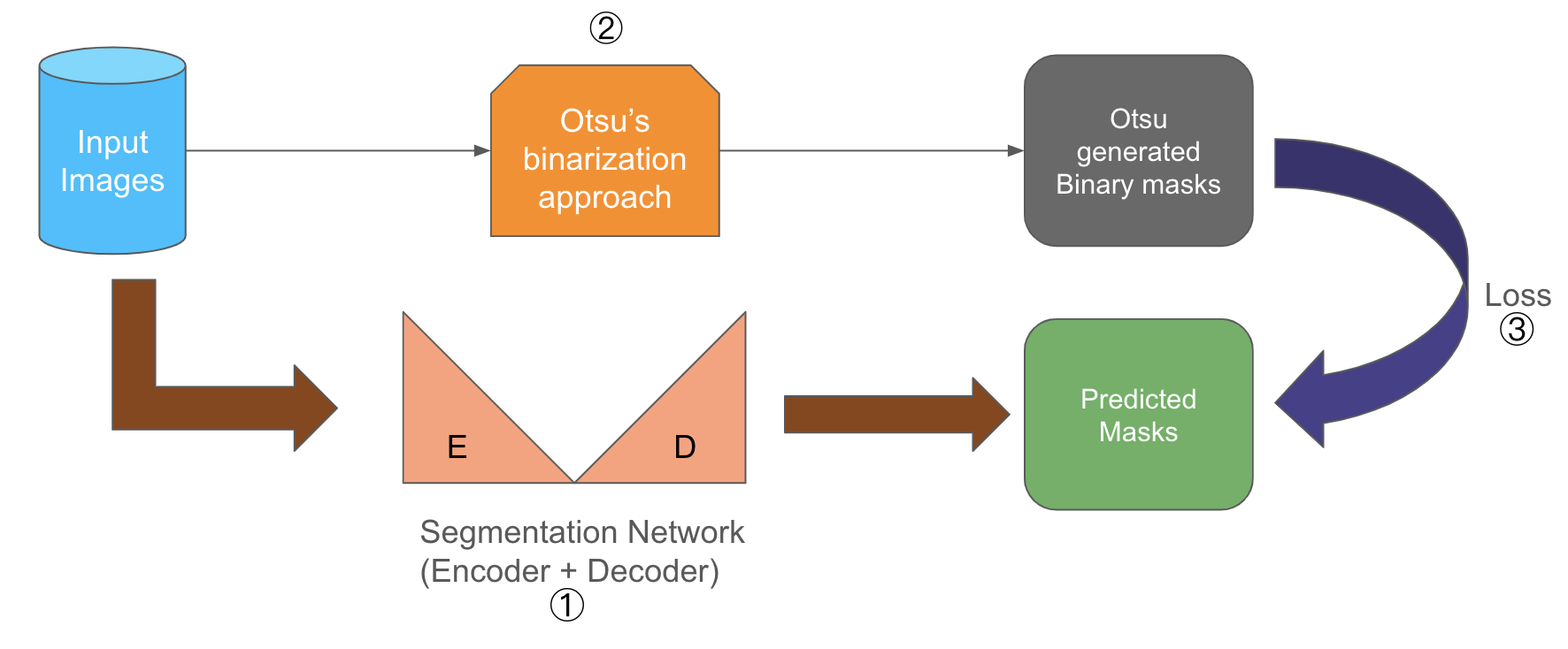}}
\vskip -0.2in
\caption{In this figure, we depict the pre-training procedure of MedSASS. Our input is only images; during each iteration, the input images are simultaneously passed through the segmentation network and Otsu's approach \cite{otsu1979threshold}. The output labels of these two are used to compute loss and optimize the segmentation network. We ablate the three fundamental components of MedSASS: the segmentation architecture (\raisebox{.5pt}{\textcircled{\raisebox{-.9pt} {1}}}), the thresholding approach for self-supervision (\raisebox{.5pt}{\textcircled{\raisebox{-.9pt} {2}}}), and  the loss function (\raisebox{.5pt}{\textcircled{\raisebox{-.9pt} {3}}}) in Sections \ref{results-all} , \ref{thresholding-cmp} and \ref{section-loss-abl} respectively.}
\label{encoder-decoder-diagram}
\end{center}
\vskip -0.2in
\end{figure}


\subsection*{Generalizable Insights about Machine Learning in the Context of Healthcare}

\begin{itemize}
    \item We start by benchmarking existing self-supervised approaches for medical image segmentation in Section \ref{ssl-bench-Section}. To the best of our knowledge, this is the first large-scale study covering multiple self-supervised techniques for CNNs and Vision Transformers on multiple challenging medical datasets.
  \item We propose a novel self-supervised technique for medical image segmentation, \textbf{M}edical imaging \textbf{E}nhanced with \textbf{D}ynamic \textbf{S}elf-\textbf{A}daptive \textbf{S}emantic \textbf{S}egmentation (MedSASS) in Section \ref{desc-of-MedSASS}. This technique improves over existing state-of-the-art self-supervised techniques by 3.83\% for CNNs and on-par for ViT-based semantic segmentation architecture over four diverse medical image segmentation datasets with encoder-only training.
  \item Additionally, unlike existing self-supervised techniques that only train an encoder, with MedSASS we can train end-to-end (an encoder as well as a decoder). With this approach, MedSASS improves segmentation performance (IoU metric) by 14.4\% for CNNs and 6\% for ViT over existing self-supervised techniques.
  
\end{itemize}

\section{Related Work}
\label{related-work}
\vskip -0.1in
Recent developments in self-supervised learning literature predominantly feature two main pre-training approaches: (i) the use of contrastive loss, and (ii) masked image modeling. Contrastive loss-based methods focus on learning through the matching of similarities and dissimilarities. Traditionally, such techniques have optimized distances between representations of different augmented views of the same image (referred to as ‘positive pairs’) and those from different images (‘negative pairs’), as demonstrated by \cite{He2020MomentumCF,Chen2020ASF,caron2020unsupervised}. However, this approach is memory-intensive, necessitating the tracking of both positive and negative pairs. Recent state-of-the-art approaches like Bootstrap Your Own Latent (BYOL) \cite{grill2020bootstrap} and Distillation with NO labels (DINO) \cite{caron2021emerging} have enhanced this performance by omitting the need for memory banks. The use of negative pairs in these models serves to prevent collapse, a challenge that has been tackled through various strategies. BYOL employs a momentum encoder, while Simple Siamese (SimSiam) \cite{Chen2021ExploringSS} utilizes a stop gradient. DINO, on the other hand, employs a combination of sharpening and centering to counterbalance and avoid collapse. Techniques that rely solely on positive pairs have proven to be more efficient than those requiring both positive and negative pairs. At the same time, there has been a rise in reconstruction-based, self-supervised pre-training methods like Simple Framework for Masked Image Modeling (SimMIM) \cite{xie2022simmim} and Masked Auto Encoders (MAE) \cite{he2022masked}. These methods aim to comprehend the semantic aspects of images by masking certain sections and subsequently predicting these masked portions during pre-training.

Notably, current self-supervised techniques primarily train only an encoder. Yet, for effective segmentation, both an encoder and a decoder equipped with robust priors are essential. While reconstructive self-supervised methods (like SimMIM \cite{xie2022simmim} and MAE \cite{he2022masked}) do involve a decoder during pre-training, they are not directly applicable for segmentation tasks. For instance, SimMIM employs a linear layer decoder, which is considerably simpler than a U-Net \cite{ronneberger2015u} decoder and focuses solely on predicting the masked portion. Concurrently, the concept of training an encoder-decoder architecture end-to-end has been investigated in fields like self-supervised depth estimation \cite{godard2019digging}. While the efficacy of existing state-of-the-art self-supervised methods has been extensively examined in the realm of natural imaging, only a limited number of studies have delved into their application in medical image segmentation. Those that do include self-supervised medical image segmentation typically have a constrained scope, with a limited range of approaches and techniques explored, as indicated in \cite{kalapos2022self}. 

To the best of our knowledge, there are no comprehensive studies that extensively explore self-supervised medical image segmentation across various medical modalities. Moreover, there is a lack of investigation into the impacts of pre-training both the encoder and decoder within the context of self-supervised medical image segmentation. As mentioned in Section \ref{introduction-start}, to the best of our knowledge this is the first extensive study of (i) benchmarking existing self-supervised approaches for medical image segmentation on multiple medical modalities and (ii) developing a novel end-to-end self-supervised medical image segmentation approach.

\section{Methods}

Our experimental approach aligns with the methodologies of \citet{caron2021emerging} and \citet{jabri2020space}, focusing exclusively on pre-trained features for segmentation. We have conducted a thorough pre-training of all self-supervised models for 50 epochs, utilizing a batch size of 16 and early stopping patience of five epochs. We observe that some of the datasets have different sizes of images in training and testing. To ensure consistency, we resize all images to $224 \times 224$ for CNN as well as ViT-based U-Net \cite{ronneberger2015u}. The CNN-based U-Net uses ResNet-50 as its backbone, while the ViT-based U-Net uses ViT-small backbone. We further descibe the ViT-based U-Net in Appendix \ref{vit_unet_desc}. Additionally, all architectures are trained from scratch, and the performance is averaged over five seed values to ensure the generalizability and statistical significance of our results. During encoder-only training, we train only the backbone and swap out any other parts for a fair comparison. Since existing state-of-the-art self-supervised training approaches only train encoders, they cannot be used to train encoders and decoders. Masked image modelling approaches like MAE and SimMIM do use a decoder, but it is only used to predict the masked part, as loss is computed only on the masked part for these approaches.

\subsection{Datasets}
\paragraph{The Dermatomyositis dataset}\cite{singh2023data, VANBUREN2022113233}
This dataset is made up of whole-slide images that were cut from muscle biopsies of patients with Dermatomyositis, stained with different proteins, and then imaged to make a dataset of 198 TIFF images from 7 patients. We use the DAPI-stained images for the semantic segmentation, in line with previous works \cite{singh2023data,VANBUREN2022113233}. Each whole slide image is 1408 by 1876. We tile these whole slide images into a size of $480 \times 480$. We then split these datasets into a ratio of 70/10/20 for training, validation, and testing, respectively. This split yields 1656, 240, and 480 images in the training, validation, and test sets, respectively. 
\paragraph{Tissuenet dataset}\cite{greenwald2022whole} This dataset contains over a million cells annotations from nine different organs, and six different imaging systems are annotated. Each cell has both nuclear and whole-cell information. Each image in the validation and test sets is $256 \times 256$, while training images are $512 \times 512$ in size. We use the official splits of the dataset for training, validation, and testing, with 2580, 3140, and 1249 images, respectively.
\paragraph{ISIC-2017 dataset} \cite{codella2017skin} The International Skin Imaging Collaboration (ISIC) 2017 image dataset is one of the largest repositories of images of melenoma skin lesions, the most lethal skin cancer. Each image has binary image masks of the lession region and background. The dataset is segmented into three parts: training, validation, and testing, with 2000, 150, and 600 images, respectively. The ISIC-2017 is a particularly hard dataset due to high intra-class variability and inter-class similarities with obscuration areas of interest \cite{Singh_2023_ICCV}.
\paragraph{X-ray image dataset}\cite{chowdhury2020can,rahman2021exploring} This is the largest dataset used in our study, with a total of 21,165 images. We split the dataset into 14,730, 2202, and 4,233 images for training, validation, and testing, respectively. The X-ray dataset contains chest X-ray images of patients with COVID-19, normal, and other lung infections. Each image in the dataset is $299 \times 299$.

Across all four datasets, our task is semantic segmentation, i.e., segmenting an area of interest (foreground) from the area of no interest (background). For evaluation on the test set, we use the IoU (Intersection over Union) score for comparison. Additionally, we further expand upon the challenges associated with each of these datasets along with samples in Appendix \ref{samples}. 

\subsection{Benchmarking existing self-supervised approaches}
\label{ssl-bench-Section}

\begin{figure}[ht]
\begin{center}
\centerline{\includegraphics[width=0.7\linewidth]{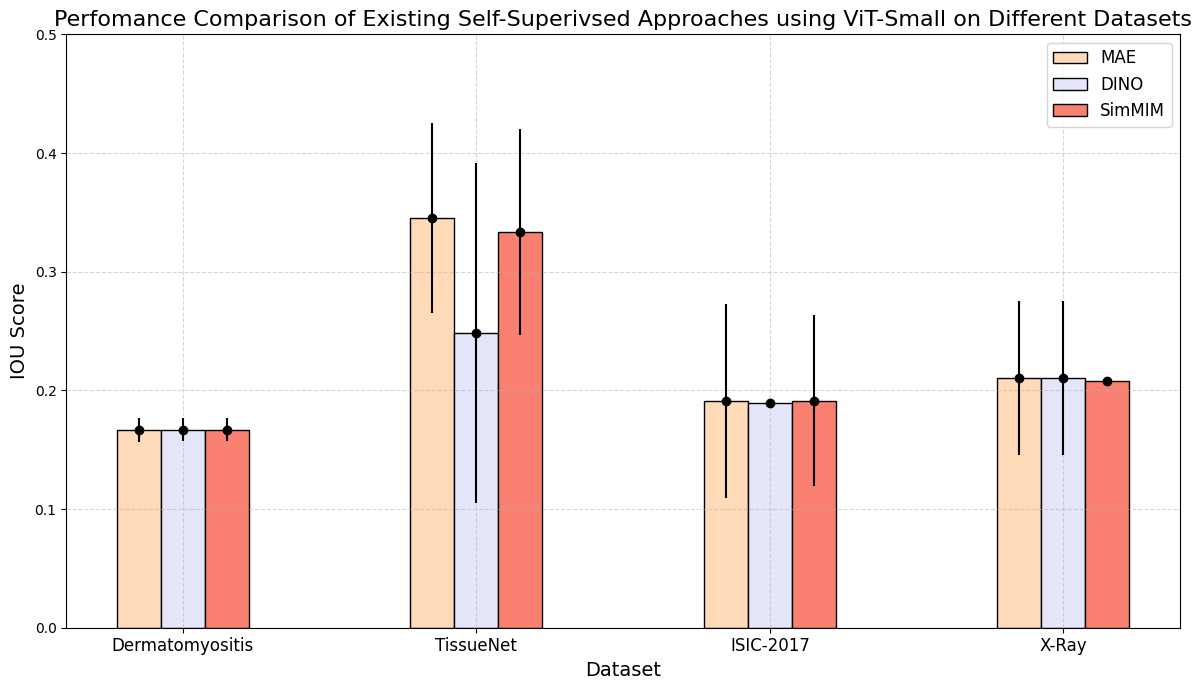}}
(a)
\centerline{\includegraphics[width=0.7\linewidth]{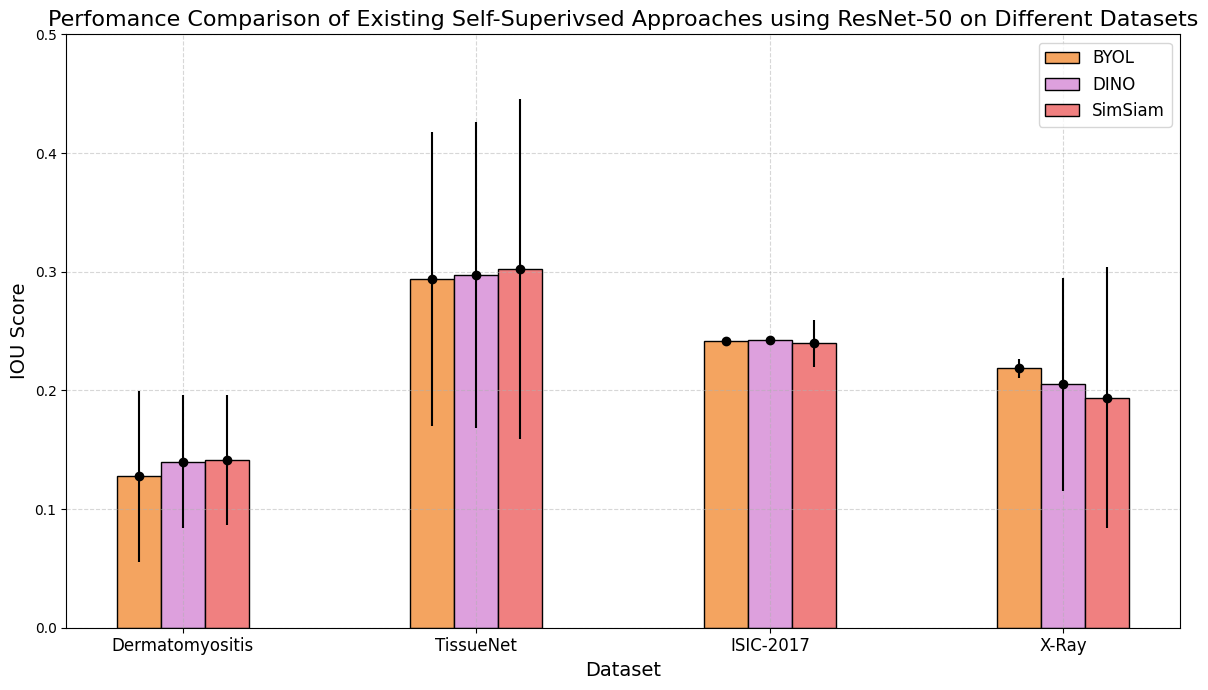}}
(b)
\vskip -0.1in
\caption{In this figure, we provide a comprehensive comparison of various pre-training strategies for Vision Transformers (sub-figure a) and Convolutional Neural Networks (CNNs) (sub-figure b). Each bar is topped with a black dot connected by a horizontal line, denoting the error margin calculated as the average across five different seed values. For Vision Transformers, our analysis reveals a uniform performance across the Dermatomyositis, X-Ray, and ISIC-2017 datasets among all evaluated methodologies. However, the TissueNet dataset distinctly showcases the superiority of masked image modeling techniques over contrastive methods such as DINO. Conversely, sub-figure (b) illustrates that there is no singularly dominant approach for CNNs, indicating the competitive nature of the current state-of-the-art methodologies. The black dot on top of each bar with horizontal line represent the error bar for average over five seed values.}
\label{ssl_bench}
\end{center}
\vskip -0.3in
\end{figure}

Our study initiates with an evaluation of diverse self-supervised learning methodologies applied to four distinct medical imaging datasets, utilizing both Convolutional Neural Networks (CNNs) and Vision Transformers (ViTs). It's noteworthy that certain approaches are limited to an architecture, for example masked image modelling approaches are only limited to Vision Transformers. We start our analysis with contrastive methods accompanied by distillation, such as BYOL and DINO, alongside contrastive techniques without momentum encoders like SimSiam, using the ResNet-50 architecture. In parallel, for ViTs, our benchmarks extend to both contrastive (e.g., DINO) and masked image modeling approaches, including SimMIM and MAE. The Intersection over Union (IoU) metrics on the test sets for both CNN and ViT-based models are depicted in Fig. \ref{ssl_bench}. Within the realm of ViTs, we observe a uniform performance across three datasets, with a notable exception being the TissueNet dataset, where masked image modeling significantly outperforms the contrastive method (DINO). Conversely, for CNNs, SimSiam slightly edges out DINO and BYOL in the Dermatomyositis and TissueNet datasets, whereas BYOL exhibits superior performance on the remaining datasets (ISIC-2017 and X-ray). Collectively, these findings underscore the absence of a universally dominant approach that consistently delivers state-of-the-art results across all datasets for CNNs.

\subsection{Intuition for MedSASS}
\label{desc-of-MedSASS}
From Section \ref{ssl-bench-Section}, we observe that for CNNs there is no single approach that is dominant across datasets. Furthermore except for masked image modelling approaches, the pre-training objective of most self-supervised approaches is aligned with classification. In this section, we propose a novel self-supervised approach with a more semantic-segmentation focused pre-training approach. For this we intorduce \textbf{M}edical imaging \textbf{E}nhanced with \textbf{D}ynamic \textbf{S}elf-\textbf{A}daptive \textbf{S}emantic \textbf{S}egmentation (MedSASS). MedSASS incorporates a semantic segmentation architecture: U-Net, selected for its preeminence in biomedical image segmentation \cite{ronneberger2015u}. During pre-training, Otsu's label are used as a psuedo-label to compute loss against the output from the U-Net architecture. The objective of the self-supervised training is to predict the Otsu's output. The choice of Otsu's method capitalizes on the intrinsic properties of input images as supervisory signals, embodying the essence of self-supervised learning. This pre-training objective is very much aligned with the supervised semantic segmentation objective wherein instead of psuedo labels we use real labels for computing loss. We present this approach schematically in Fig. \ref{encoder-decoder-diagram}. From Fig. \ref{encoder-decoder-diagram}, we observe that both the encoder and the decoder can be trained using this approach as opposed to the existing encoder only self-supervised approaches. Since, existing state-of-art approaches only train an encoder we start by training MedSASS in an enoder only fashion followed by end-to-end (encoder and decoder) training to study if pre-training the decoder architecture improves performance. We start by explaining the MedSASS pre-training objective in Section \ref{pretrainng-objective}, followed by explaining how we are avoiding collapse in MedSASS in Section \ref{collapse-rationale}. 

\subsubsection{MedSASS' Pre-training Objective}
\label{pretrainng-objective}
As elucidated in Fig. \ref{encoder-decoder-diagram}, the MedSASS self-supervised pre-training objective is very similar to a supervised semantic segmentation objective. In MedSASS for each batch of input images, we pass them through \cite{otsu1979threshold}'s algorithm to obtain binarized outputs. These binarized outputs are then used as psudo labels to compute loss with the semantic masks predicted by the U-Net. Since, the Otsu's appraoch uses intrinsic properties of input images as supervisory signals, there is no need for an actual ground truth label to supervise the training, hence the pre-training approach is self-reliant and self-supervised in nature. The quality of masks generated by Otsu's is also important for effective pre-training. We discuss other approaches to obtain binarized labels in Section \ref{why-otsu}.

\subsubsection{Why \cite{otsu1979threshold}'s approach for Psuedo-labels?}
\label{why-otsu}

\begin{table*}[]
\centering
\begin{tabular}{l|l|l|l|l|l}
\hline
Approach   & Backbone  & \multicolumn{4}{c}{Datasets}                                                                                                                                                                                                                                                \\
           &           & DM                                                               & TissueNet                                                         & ISIC-2017                                                         & X-Ray                                                            \\
           \hline
DINO       &ResNet-50 & \begin{tabular}[c]{@{}l@{}}0.1399\\ ±0.056\end{tabular}          & \begin{tabular}[c]{@{}l@{}}0.2972\\ ±0.129\end{tabular}           & \begin{tabular}[c]{@{}l@{}}0.2426\\ ±0.002\end{tabular}           & \begin{tabular}[c]{@{}l@{}}0.2052\\ ±0.089\end{tabular}         \\
BYOL       & ResNet-50 & \begin{tabular}[c]{@{}l@{}}0.1278\\ ±0.072\end{tabular}          & \begin{tabular}[c]{@{}l@{}}±0.3455\\ ±0.08\end{tabular}           & \begin{tabular}[c]{@{}l@{}}0.2412\\ ±0.002\end{tabular}          & \begin{tabular}[c]{@{}l@{}}0.2185\\ ±0.008\end{tabular}          \\
SimSiam    & ResNet-50 & \begin{tabular}[c]{@{}l@{}}0.1415\\ ±0.054\end{tabular}          & \begin{tabular}[c]{@{}l@{}}0.3021\\ ±0.143\end{tabular}          & \begin{tabular}[c]{@{}l@{}}0.2395\\ ±0.019\end{tabular}           & \begin{tabular}[c]{@{}l@{}}0.1939\\ ±0.110\end{tabular}          \\
MedSASS(e) & ResNet-50 &\cellcolor[HTML]{34FF34} \begin{tabular}[c]{@{}l@{}}0.1742\\ ±0.048\end{tabular} &\cellcolor[HTML]{34FF34}\begin{tabular}[c]{@{}l@{}}0.3824\\  ±0.005\end{tabular} &\cellcolor[HTML]{34FF34}\begin{tabular}[c]{@{}l@{}}0.2445\\  ±0.004\end{tabular} & \cellcolor[HTML]{34FF34}\begin{tabular}[c]{@{}l@{}}0.2568\\ ±0.016\end{tabular} \\
\hline
\end{tabular}
\vskip -0.1in
\caption{The table showcases the Intersection over Union (IoU) scores averaged across five distinct seed values for a U-Net architecture equipped with a ResNet-50 backbone, tested across four diverse medical imaging datasets. Here, DM represents Dermatomyositis. To highlight the superior performance, the top-performing numbers are emphasized in \fcolorbox{black}{mygreen}{\makebox(20,10){}} for each dataset. MedSASS (e) is indicative of our focus on training only the encoder of MedSASS, aligning with the training scope of other self-supervised techniques for a balanced comparison. Notably, MedSASS demonstrates superior performance over other leading self-supervised methods across all datasets, employing a ResNet-50 backbone. This observation underscores the effectiveness of MedSASS in enhancing the capabilities of self-supervised learning in medical image analysis. }
\label{cnn-encoder-only}

\end{table*}

MedSASS aims to pioneer a self-supervised approach for medical semantic segmentation. Typically,semantic segmentation in medical imaging entails differentiating the area of interest (foreground) from the non-relevant area (background). For instance, in skin lesion imaging, the task involves isolating the lesion (foreground) from the rest of the dermatology slide, and similarly, in histopathology slides, distinguishing cells (foreground) from the slide background. Consequently, this necessitates binary masks.

In MedSASS, we employ Otsu's approach, a classic technique widely used in medical imaging, for semantic segmentation to this day on datasets that are too small to train using deep neural networks \cite{barron2020generalization}. During pre-training the architecture is trained to predict masks generated by Otsu's approach. There are certain priors associated with medical imaging that make the application of Otsu's approach favorable in medical images as compared to natural images. In this section we provide the rationale for using Otsu's approach, followed by empirical results in Section \ref{results-all} and qualitative samples in Appendix \ref{samples}. These priors include:
\begin{itemize}
\item In medical imaging, high contrast is ensured to best visualize disease features. Otsu's approach effectively separates high-contrast areas, making it ideal for medical images \cite{dance2014diagnostic}.

\item Medical images typically have a more straightforward and consistent background-foreground relationship. For instance, in an X-ray, the bones might be the foreground, and everything else is the background. Otsu's approach, which works by finding a threshold that best separates these two classes, is well-suited for such scenarios. Natural images, however, often contain multiple objects, textures, and gradients, making it difficult for a single threshold to effectively segment all relevant objects as elucidated in Appendix \ref{natural-datasets}.  In medical imaging, the primary goal of segmentation is often to isolate specific anatomical structures or pathologies, which are usually distinct in intensity from the surrounding tissue. Otsu's approach efficiently accomplishes this by focusing on the intensity differences. In natural imaging, segmentation goals are more diverse and may require identifying and separating multiple objects and textures, which necessitates more sophisticated segmentation techniques. We further empirically validate these in Appendix \ref{natural-datasets} where we observe that MedSASS with Otsu's approach performs sub-par as compared to other self-supervised techniques. 

\item Medical imaging techniques are usually conducted in controlled environments with uniform lighting, which leads to more consistent image quality. This consistency is conducive to the application of Otsu's technique, which assumes a bimodal histogram (two distinct peaks representing the foreground and background). On the other hand, natural images are taken under varying lighting conditions and may not exhibit a clear bimodal histogram, reducing the effectiveness of Otsu's approach \cite{pham2000current}.
\end{itemize}

Until now Otsu's masks have been used for comparison against neural networks. To the best of our knowledge, MedSASS is the first approach to leverage this ``automatic" approach for self-supervision \cite{kumar2017dataset}. In our methodology, we first convert images to grayscale and then apply Otsu's technique to generate thresholded masks. Otsu's technique dynamically calculates the threshold for each image, enabling the segmentation architecture to learn meaningful representations rather than adapting to a specific threshold value. We also juxtapose MedSASS' performance using other ``automatic" methods for obtaining binarized masks, such as adaptive thresholding \citet{661196} and Generalized Histogram Thresholding (GHT) \cite{barron2020generalization}, to underscore the efficacy of Otsu's approach in diverse imaging scenarios in Section \ref{thresholding-cmp}. In addition to this we provide samples of Otsu generated binary masks in Appendix \ref{samples}. We observe that despite the complex nature of input images, Otsu's output is very similar to the ground truth mask.
\subsubsection{How MedSASS avoids collapse ?} 
\label{collapse-rationale}

In MedSASS, we circumvent collapse through two pivotal strategies: firstly, by ensuring that the architecture eschews trivial learning to foster learning meaningful priors, and secondly, by promoting equitable learning across both majority and minority classes. In this section, we theoretically explain these concepts followed by quantitative results in Section \ref{thresholding-cmp} and Section \ref{section-loss-abl} respectively as an ablation study. 

The efficacy of the generated masks is crucial for MedSASS, as it depends on them for supervision. The simplest method to create these masks involves setting a fixed threshold value for supervision. However, this approach has a significant drawback: the model might not learn meaningful features but rather categorize pixels based solely on whether they fall above or below this threshold. This results in the architecture merely learning to threshold at the set value instead of acquiring useful features. Moreover, applying a uniform global threshold may not be effective, as different areas within an image could exhibit varying pixel intensities. In such cases, a global threshold might obscure details that are critical for the architecture to learn valuable representations. Furthermore, this could also lead to dimensional collapse, wherein instead of learning useful latent representation, they are compressed and clustered at a single point.

To address this challenge, adaptive thresholding can be employed. This technique adjusts the threshold for each pixel based on the surrounding region, providing variable thresholds for different image areas, which is advantageous in images with diverse pixel intensities. Nonetheless, adaptive thresholding methods also necessitate an initial arbitrary pixel value. To circumvent the need for a predefined threshold, techniques like Otsu's approach \cite{otsu1979threshold} can be utilized. 
In Otsu's approach, each image has a different threshold value determined by minimizing intra-class intensity variance, or equivalently, by maximizing inter-class variance. Thus, a pixel with the same pixel intensity can be classified as background or foreground based on its context, as opposed to the case when we simply set a threshold and all values are clustered based on their values, irrespective of the context. Setting a fixed threshold \( x \) in image segmentation implies that all pixel values below \( x \) are classified as background (bg) across the entire dataset, regardless of their context. This results in their representations converging to a singular point in the latent space. However, in Otsu's approach, the threshold is not constant but varies from one image to another. Consequently, while pixels below a certain value \( x_1 \) might be classified as background in one image, they could be deemed foreground (fg) in another image within the same dataset. This variability compels the architecture to genuinely learn and discern useful image features, enhancing its ability to make accurate prior predictions and avoid dimensional collapse. Since, the thresholds are determined on an image level, Otsu's approach is "self-adaptive" in nature. Otsu's method, widely used in medical imaging and OCR, is particularly beneficial for small datasets that are challenging to train. Its popularity has led to the development of an improved and generalized version combinig Otsu's method with minimum error thresholding \cite{kittler1986minimum} into a Generalized Histogram Thresholding \cite{barron2020generalization}. We qualitatively study the effect of altering the thresholding approach in Section \ref{thresholding-function-abl}.


But even with this non-trivial thresholding, there is a possibility that the image is split into an imbalanced number of foreground and background pixels. And during training, the architecture overfits to one of them. This does happen when we use cross-entropy loss, as elucidated in \cite{lin2017focal,Singh_2023_ICCV} and Section \ref{loss-function-abl}. To prevent this from happening, we use focal-tversky loss function. 

\subsubsection{Why focal-tversky loss ?} 
\label{focal-loss-rationale}
Region- or distribution-based loss functions are commonly utilized in medical image segmentation. For instance, binary cross-entropy loss, a variant of cross-entropy loss, exemplifies a distribution-based loss function. However, it often tends to overfit the predominant class in binary image segmentation \cite{Singh_2023_ICCV}, which can be either the foreground or background. To address this issue, focal loss \cite{lin2017focal} was developed. It differentially weights the majority and minority classes, thereby preventing overfitting of the majority class through targeted penalization.

In contrast, DICE loss \cite{sudre2017generalised}, a region-based loss function, is adapted from the DICE coefficient used in evaluating segmentation performance. However, it equally treats False Positives (FP) and False Negatives (FN), which is a notable limitation in medical imaging. The impact of FPs and FNs in this field is asymmetric; for example, missing a pathological feature (FN) can be more detrimental than a false alarm (FP), as it could result in missed diagnoses and delayed treatments. To mitigate this, tversky loss \cite{salehi2017tversky} was introduced, applying variable weights to the FP and FN classes. Our selection of focal-tversky \cite{abraham2019novel} loss is driven by its hybrid nature, combining elements of both distribution and region-based loss functions. This approach effectively prevents overfitting to a specific pixel class and provides a focused addressal for FNs, aligning more closely with the nuanced needs of medical image analysis. We present these results in Section \ref{thresholding-cmp} and Table \ref{loss-function-abl}. From this table, we observe that focal-tversky loss performs much better than focal and tversky loss.

\label{experimental-setup}

\begin{table*}[ht]
\centering
\begin{tabular}{l|l|l|l|l|l}
\hline
Approach   & Backbone & \multicolumn{4}{c}{Datasets}                                                                                                                                                                                                             \\
           &          & DM                                                       & TissueNet                                                & ISIC-2017                                                & X-Ray                                                   \\
           \hline
DINO       & ViT-small    & \begin{tabular}[c]{@{}l@{}}0.1668 \\ ±0.008\end{tabular} & \begin{tabular}[c]{@{}l@{}}0.2483 \\ ±0.143\end{tabular} & \begin{tabular}[c]{@{}l@{}}0.1891\\ ±0.0031\end{tabular} & \begin{tabular}[c]{@{}l@{}}0.2106\\ ±0.065\end{tabular} \\

MAE        & ViT-small    & \begin{tabular}[c]{@{}l@{}}0.1668\\  ±0.009\end{tabular} & \begin{tabular}[c]{@{}l@{}}0.3455\\ ±0.08\end{tabular}  & \begin{tabular}[c]{@{}l@{}}0.191\\ ±0.081\end{tabular}   & \begin{tabular}[c]{@{}l@{}}0.2106\\ ±0.064\end{tabular} \\

SimMIM     & ViT-small    & \cellcolor[HTML]{34FF34}\begin{tabular}[c]{@{}l@{}}0.167\\ ±0.009\end{tabular}   & \begin{tabular}[c]{@{}l@{}}0.3335\\ ±0.086\end{tabular}  & \begin{tabular}[c]{@{}l@{}}0.1913\\±0.072\end{tabular}  & \begin{tabular}[c]{@{}l@{}}0.2077\\ ±0.002\end{tabular} \\

MedSASS(e) & ViT-small    & \begin{tabular}[c]{@{}l@{}}0.136\\ ±0.079\end{tabular}   & \cellcolor[HTML]{34FF34} \begin{tabular}[c]{@{}l@{}}0.3633\\±0.006\end{tabular}  & \cellcolor[HTML]{34FF34} \begin{tabular}[c]{@{}l@{}}0.1914\\ ±0.083\end{tabular}  & \cellcolor[HTML]{34FF34} \begin{tabular}[c]{@{}l@{}}0.2466\\ 
±0.003\end{tabular}\\
\hline
\end{tabular}
\vskip -0.1in
\caption{In this table, we present the results of U-Net trained with ViT-small backbone on four diverse medical imaging datasets over five different seed values. Numbers mentioned in \fcolorbox{black}{mygreen}{\makebox(20,10){}} colour are 
the best performing approach for each dataset. Here, DM represents Dermatomyositis.
Additionally, MedSASS (e) represents that we are only training the encoder of MedSASS, for fair comparison with other self-supervised techniques.  We observe that MedSASS outperforms existing state-of-the-art self-supervised approaches for three out of the four datasets with a ViT-small backbone. Furthermore, We observed that masked image modeling-based self-supervised techniques like MAE and SimMIM perform better than contrastive self-distilling approaches like DINO.}
\label{Tab:vit-s-encoder-only}
\end{table*}

\vskip -0.4in
\section{Results}
\label{results-all}

\subsection{Encoder only training}
\label{encoder-only}

Existing self-supervised techniques pre-train an encoder that can then be used on a variety of downstream tasks. We start by only training the encoder of MedSASS. During this encoder-only pre-training, we only keep the encoder and discard everything else. We then transfer the saved weights of this encoder to evaluate the testing set of each dataset. We don't fine-tune and use the learned features for semantic mask prediction, similar to \cite{caron2021emerging} and \cite{jabri2020space}. We present the results of this training procedure with a ResNet-50 backbone in Table \ref{cnn-encoder-only} and with a ViT-small backbone in Table \ref{Tab:vit-s-encoder-only}. We observed that MedSASS outperforms existing state-of-the-art self-supervised techniques with a CNN backbone, U-Net. For the ViT-small U-Net, we observed that MedSASS outperforms existing state-of-the-art self-supervised techniques on three (out of four) datasets and overall is on par with them.

\subsection{Encoder with Decoder training}
\label{enc-decoder-results}
\begin{figure*}[ht]
\begin{center}
\centerline{\includegraphics[width=0.8\linewidth]{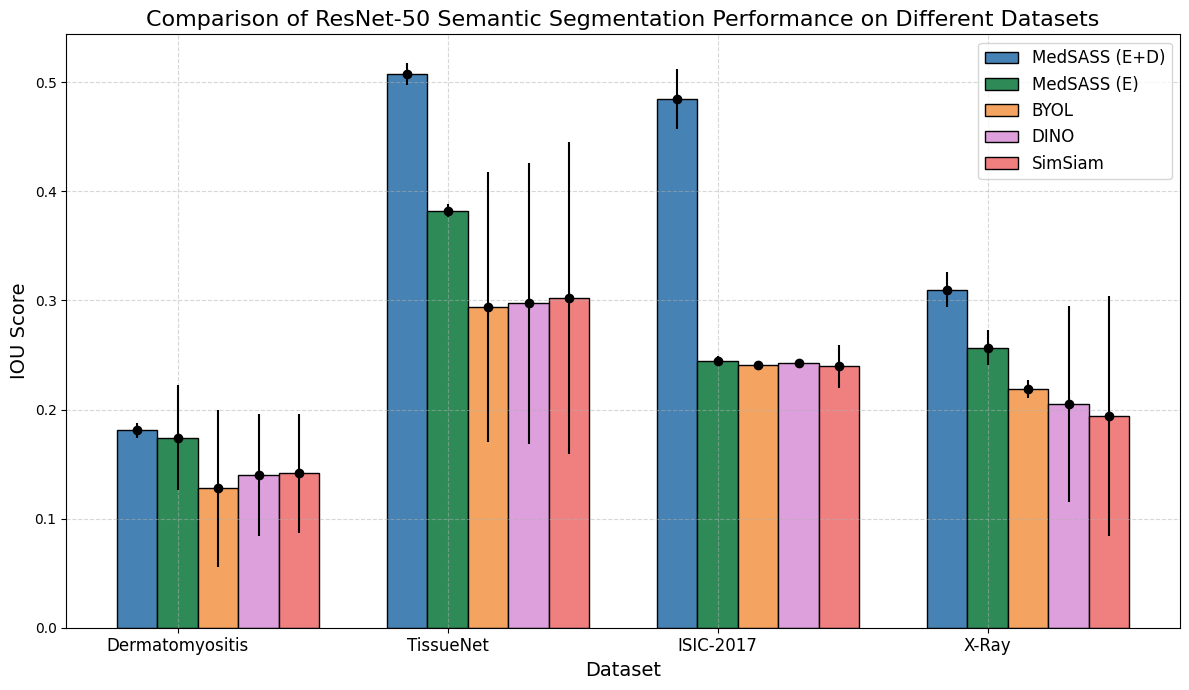}}
\caption{In this figure, we present a comparative analysis of various training methodologies: MedSASS with both encoder and decoder (denoted as MedSASS (E+D)), MedSASS utilizing only the encoder (denoted as MedSASS (E)), BYOL, DINO, and SimSiam. Our findings reveal that MedSASS, when trained solely with an encoder (indicated in green), surpasses all existing state-of-the-art self-supervised techniques in semantic segmentation performance. Moreover, the end-to-end training of MedSASS demonstrates an even more pronounced advantage over these state-of-the-art self-supervised approaches in the same domain. While the encoder only pre-trained MedSASS surpasses current state-of-the-art self-supervised methods by an average of 3.83\%, averaged over four datasets and five different seeds per dataset. Remarkably, when MedSASS is pre-trained end-to-end, it outperforms these state-of-the-art self-supervised methods by an even more significant margin of 14.4\%, also averaged over the same four datasets and seed settings. The black dot on top of each bar with horizontal line represent the error bar for average over five seed values.}
\label{encoder-decoder-cnn}
\end{center}
\vskip -0.3in
\end{figure*}

\begin{figure*}[ht]
\begin{center}
\centerline{\includegraphics[width=0.8\linewidth]{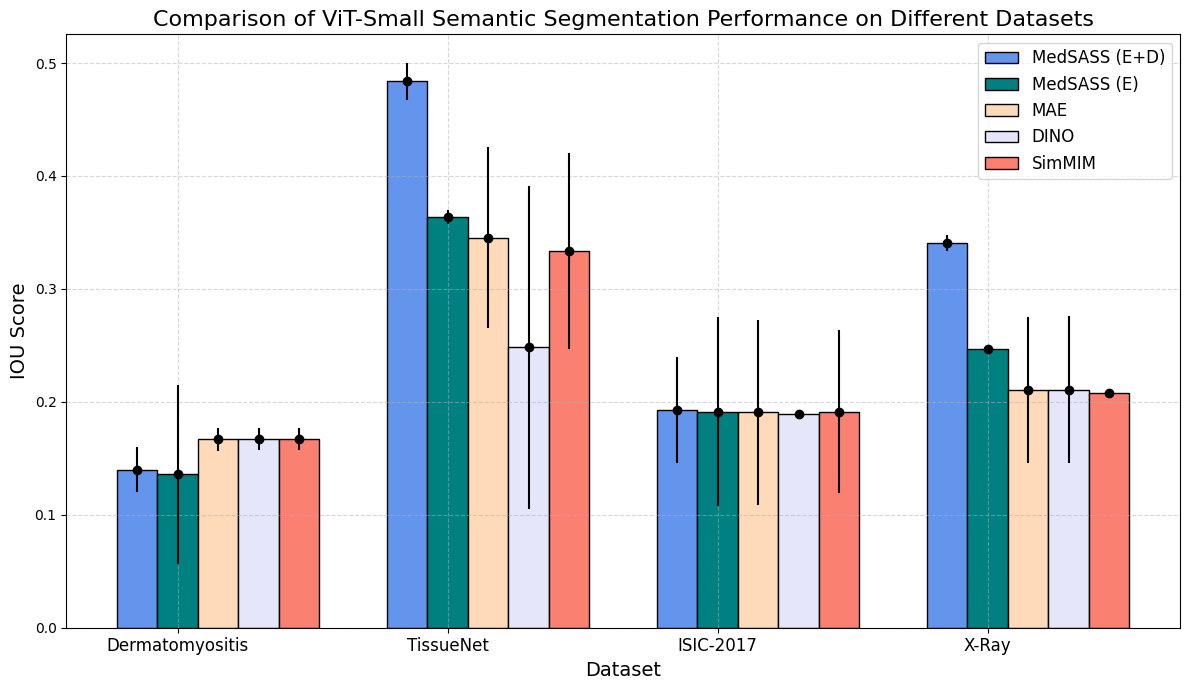}}
\vskip -0.1in
\caption{This bar graph illustrates a comparison between end-to-end and encoder-only pretrained MedSASS against other state-of-the-art self-supervised techniques, all utilizing a ViT-small backbone. We note that MedSASS, when solely pretrained, aligns closely in performance with other state-of-the-art self-supervised methods for semantic segmentation. Notably, MedSASS achieves an average improvement of 6\% over these state-of-the-art techniques when trained end-to-end, as evidenced across the four datasets. Similar to Figure \ref{encoder-decoder-cnn}, black dot on top of each bar with horizontal line represent the error bar.}
\label{encoder-decoder-vits}
\end{center}
\vskip -0.2in
\end{figure*}

As mentioned already, existing state-of-the-art self-supervised techniques aim to pretrain an encoder and then use it for downstream tasks. For classification, this is pretty straightforward, as we only need to train a small MLP on top of the pre-trained encoder. But for segmentation, usually a more complex and much deeper decoder is used to predict the mask from the latent representation extracted from the pre-trained encoder. With MedSASS, we can pre-train a U-Net end-to-end for semantic segmentation in a self-supervised way. Instead of getting rid of everything except the encoder (as mentioned in encoder-only training in Section \ref{encoder-only}), for end-to-end training, train a reusable decoder that can then be used directly for segmentation. We provide a complete set of results comparing the end-to-end encoder-only and existing state-of-the-art pre-training performance of MedSASS for CNN in Figure \ref{encoder-decoder-cnn} and in Figure \ref{encoder-decoder-vits} for ViT-small.

\section{Ablation study}
\subsection{Change in Thresholding Technique}
\label{thresholding-cmp}
As explained in Section \ref{collapse-rationale}, the efficacy of thresholding approach is important for supervision of MedSASS. In this section, we explore the impact of different thresholding techniques. We start with two examples of adaptive thresholding \cite{661196} —mean and gaussian thresholding—followed by Otsu's \cite{otsu1979threshold} approach and the more advanced, generalized Generalized Histogram Thresholding (GHT) \cite{barron2020generalization}. We train a Resnet-50 \cite{he2015deep} MedSASS end-to-end for 50 epochs with a batch size of 16 and report the IoU averaged over five seeds for four datasets in Table \ref{thresholding-function-abl}. Our findings indicate that while all techniques perform comparably on the Dermatomyositis dataset, they fall short of the performance achieved by Otsu's approach-supervised MedSASS across datasets.

\subsection{Change in Loss Function}
\label{section-loss-abl}
As mentioned in Section \ref{focal-loss-rationale}, by default, we use focal-tversky loss in MedSASS. In this section, we empirically study the effect of training the same MedSASS setup with different loss functions as mentioned in Section \ref{focal-loss-rationale}. 
We provide these results in Table \ref{loss-function-abl}. From Table \ref{loss-function-abl}, we observe that the combination of focal-tversky loss consistently performs better than either of the loss functions individually.

Furthermore, we also provide ablation study of changing the number of pre-training epochs and batch size for MedSASS in Appendix \ref{eph-bs-section}. From Table \ref{epcohs-abl}, we observed that both the CNN and ViT based MedSASS are robust to change in pre-training epoch. Similarly, for the change in batch size, from Tables \ref{bs-abl-cnn} and \ref{bs-abl-vit}, we observe that overall the performance remains constant except for a minute improvement when we switch to a batch size of 8. In addition to these, we also compare the performance of self-supervised approaches (including MedSASS) with supervised approach in Section \ref{supervised-smp-app} for both CNN as well as ViTs.

\begin{table}[]
\begin{tabular}{lllll}
\toprule
\textbf{Dataset} & \textbf{\begin{tabular}[c]{@{}l@{}}AMT\end{tabular}} & \textbf{\begin{tabular}[c]{@{}l@{}}AGT\end{tabular}} & \textbf{\begin{tabular}[c]{@{}l@{}}GHT \\ \cite{barron2020generalization}\end{tabular}}
& \textbf{\begin{tabular}[c]{@{}l@{}}Otsu \\ \cite{otsu1979threshold}\end{tabular}} \\
\midrule
DM            & 0.1724±0.002                                                                     & 0.1725 ±0.004                                                                        & 0.1809 ±0.0227  & \textbf{0.181 ±0.0068}   \\
ISIC-2017        & 0.2393 ±0.001                                                                    & 0.2393 ±0.006                                                                        & 0.1236 ±0.0039  & \textbf{0.4847±0.0277}   \\
TissueNet               & 0.3949 ±0.0087                                                                   & 0.391 ±0.00899                                                                       & 0.4927 ±0.00885 & \textbf{0.5077 ±0.0103}  \\
X-Ray            & 0.1839 ±0.00189                                                                  & 0.2079 ±0.0048                                                                       & 0.1689 ±0.0281  & \textbf{0.3522 ±0.00238} \\
\bottomrule
\end{tabular}
\vskip -0.1in
\caption{This table displays IoU scores for MedSASS using various thresholding methods: Adaptive Mean Thresholding (AMT), Adaptive Gaussian Thresholding (AGT) \cite{661196}, GHT \cite{barron2020generalization}, and Otsu's approach \cite{otsu1979threshold}, across four datasets - DM (Dermatomyositis), ISIC-2017, TissuNet and X-Ray. The results are presented as IoU scores averaged over five seed values with standard deviations, showcasing the performance variation of each technique in medical image segmentation. We observe that overall Otsu's approach provides the best supervision for MedSASS.}
\label{thresholding-function-abl}
\vskip -0.1in
\end{table}

\begin{table}[h]
\centering
\begin{tabular}{l|l|l|l}
\toprule
Dataset   & \multicolumn{3}{c}{IoU on Test Set}                                                                                                                                                            \\
          & Tversky Loss                                                & Focal Loss                                                 & Focal Tversky                                                       \\
          \hline
Derm.     & \begin{tabular}[c]{@{}l@{}}0.1726 \\ ±0.000133\end{tabular} & \begin{tabular}[c]{@{}l@{}}0.02\\ ±0.0181\end{tabular}     & \textbf{\begin{tabular}[c]{@{}l@{}}0.181 \\ ±0.0068\end{tabular}}   \\
\hline
ISIC-2017 & \begin{tabular}[c]{@{}l@{}}0.2444 \\ ±0.00597\end{tabular}  & \begin{tabular}[c]{@{}l@{}}0.05291 \\ ±0.0907\end{tabular} & \textbf{\begin{tabular}[c]{@{}l@{}}0.4847\\ ±0.0277\end{tabular}}   \\
\hline
TN.       & \begin{tabular}[c]{@{}l@{}}0.3604 \\ ±0.0301\end{tabular}   & \begin{tabular}[c]{@{}l@{}}0.11 \\ ±0.0117\end{tabular}    & \textbf{\begin{tabular}[c]{@{}l@{}}0.5077 \\ ±0.0103\end{tabular}}  \\
\hline
X-Ray     & \begin{tabular}[c]{@{}l@{}}0.1963 \\ ±0.113\end{tabular}    & \begin{tabular}[c]{@{}l@{}}0.02\\ ±0.0261\end{tabular}     & \textbf{\begin{tabular}[c]{@{}l@{}}0.3522 \\ ±0.00238\end{tabular}}\\
\bottomrule
\end{tabular}
\vskip -0.1in
\caption{This table presents a comparative analysis of MedSASS trained using various loss functions: tversky loss, focal loss, and focal-tversky loss. Additionally, we also studied the effect of using cross-entropy loss in Appendix \ref{ce-ablation-expl}. In the case of cross-entropy loss, we note a collapse in the model's performance. Further, our observations highlight that focal-tversky loss significantly outperforms the models trained with either focal or Tversky loss alone.}
\label{loss-function-abl}
\end{table}

\section{Discussion and Conclusion}

\paragraph{Impact in Healthcare}\label{impact_hlt}
Medical image analysis is one of the most pivotal and challenging domains for application of artificial intelligence. Its effectiveness hinges largely on the availability of high-quality and abundant training labels. However, acquiring these labels is costly and time-consuming. During crises, the challenge is exacerbated; clinicians and experts, who are essential for labeling images, are overwhelmingly occupied with patient care and triage. In these critical times, AI systems have the potential to alleviate the burden on clinicians through automated analysis. Self-supervised methods like MedSASS are particularly valuable in addressing this dilemma. Due to their self-supervised nature they can be scaled without labels. This attribute is especially beneficial in scenarios like region-specific diseases in developing countries, where limited resources or a shortage of experts constrain label availability. Self-supervised learning techniques in medical imaging thus offer significant advantages in such contexts. Our research has been conducted on a single GPU, testing the effectiveness of MedSASS with various pre-training epochs and batch sizes (Appendix \ref{additional-abl-section}). This approach ensures that MedSASS is both accessible and replicable for practitioners, regardless of their resource availability.\footnote{We would open-source code for MedSASS upon acceptance.}
\paragraph{Limitations} \label{limitations}MedSASS fundamentally depends on Otsu's approach for self-supervision, as elucidated in Section \ref{why-otsu}. Hence, it suffers from the same limitations as Otsu's approach. This limits the applicability of MedSASS in the context of natural images. Additionally, iterative Otsu thresholding is capable of performing multi-class semantic segmentation, but it becomes very computationally intensive. Furthermore, since Otsu's method requires conversion to grayscale, a great deal of image information is lost, and masks become worse with every iteration. This limits its transferability to other dense prediction tasks. Having said that, binary semantic segmentation is the most popular task in the medical imaging community \cite{litjens2017survey}. To assess its adaptability, we trained MedSASS on a natural imaging dataset and present the results in Appendix \ref{natural-datasets}. Furthermore, the classification performance of MedSASS is not state-of-the-art in all cases, as expanded in Appendix \ref{classification-app}.

\paragraph{Ethical Aspect}
It is important to note that all datasets utilized in our experiments are designated solely for academic and research purposes. Deploying these methodologies in real-world scenarios necessitates approval from relevant health and safety regulatory authorities. Additionally, a lot of the dermatology datasets may have certain biases in their collection and may have risk-biased performance \cite{kleinberg2022racial}.

\paragraph{Conclusion}In this paper, we introduced MedSASS, a novel end-to-end self-supervised technique for medical image segmentation. Demonstrating a notable improvement over traditional methods, MedSASS enhances semantic segmentation performance for CNN by 3.83\% and matches ViTs, based on IoU evaluations averaged across four diverse medical datasets. With end-to-end training, the average gain over four datasets increases to 14.4\% for CNNs and 6\% for ViT-small. Since, MedSASS is self-supervised and can be scaled without labels. It offers promising prospects for bolstering healthcare solutions, especially in settings marred by scarce label availability, such as in the treatment of region-specific diseases in underdeveloped nations or amidst crises that demand medical professionals' undivided attention towards patient care and triage.
\newpage
\bibliography{sample}

\newpage
\appendix
\section{Additional Implementation Details}
\label{additional-implementation-result}
\algrenewcommand\algorithmicrequire{\textbf{Input:}}
\algrenewcommand\algorithmicensure{\textbf{Output:}}

\begin{algorithm}[tb]
 \label{alg:algorithm}
   \caption{MedSASS self-supervised algorithm}
   \begin{algorithmic}[1]
  \Require Unlabeled same augmented images from the training set $x$
\State $Output_{model} = Unet(x)$ \Comment{Output from the untrained Unet}
\State $Label_{OTSU} = OTSU_{openCV}(x)$ \Comment{Creating Pseudo-labels using OTSU's appraoch}
\State $loss = focal\_tversky(Output_{model},Label_{OTSU})$ \Comment{Calculating Focal-Tversky Loss}
\State Backpropogate loss.
\end{algorithmic}
\label{algorithm}
\end{algorithm}

In Algorithm \ref{alg:algorithm}, $Unet()$ represents the encoder-decoder architecture, we aim to train. $Output_{model}$ is the output we we get after passing in the input image through the segmentation architecture, in this case - $Unet()$. Finally, we obtain pseudo-labels from Otsu's approach \cite{otsu1979threshold} and compare the focal-tversky loss between Otsu's output and model's output to train the $Unet()$. 

\subsection{Common Implementation}

We use the ResNet-50-based U-Net as used in \cite{singh2023data}. This U-Net consists of an encoder and decoder depth of three with channel ranges of 256, 128, and 64. The decoder also contains channel-level attention. For implementing the CNN-based U-Net, we use \cite{Iakubovskii:2019} and all of our code has been implemented in PyTorch \cite{paszke2019pytorch}.

\begin{figure*}[ht]
\begin{center}
\centerline{\includegraphics[width=0.8\linewidth]{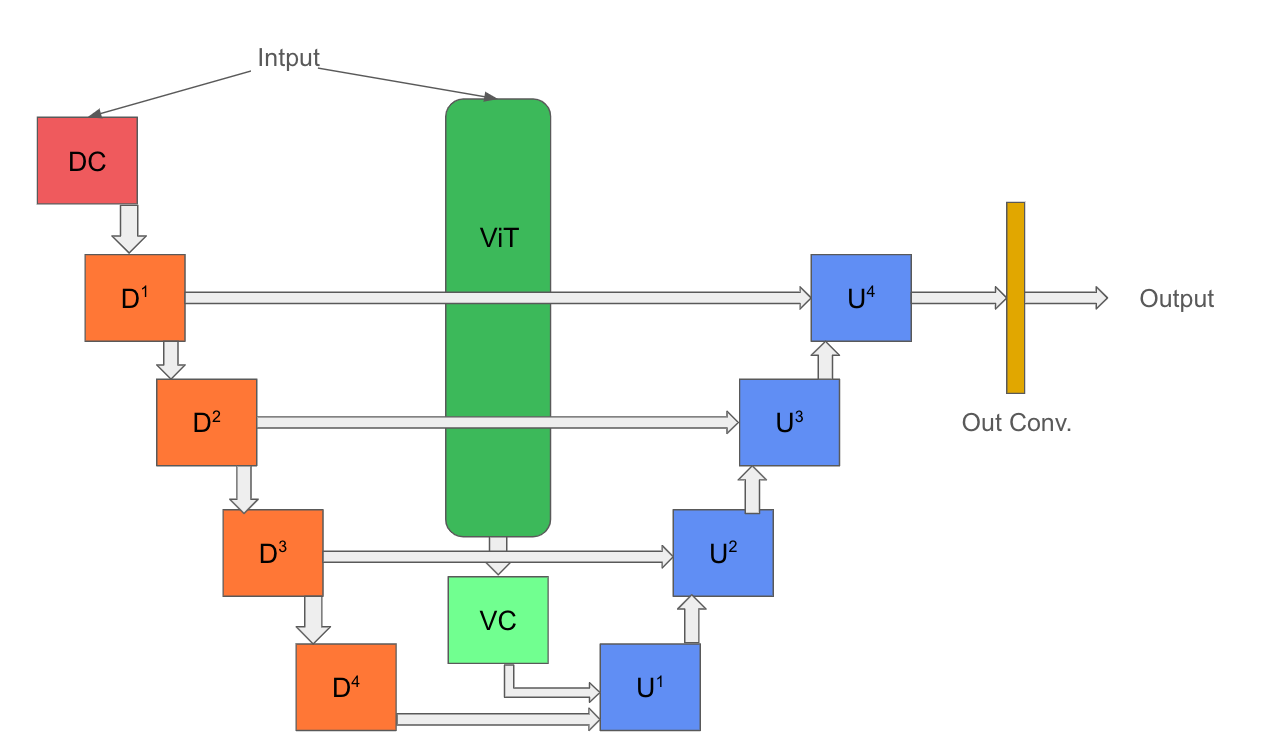}}
\caption{This figure provides a comprehensive view of the ViT-small based U-Net architecture used in our study.In this figure above, DC represents Double Convolution block, $DC^i$ represents the $i$th Downsampling block used, $U^j$ represents the $j$th upsampling block used. Finally, ViT here represents the Vision-transformer small encoder used and VC represents the Vision Transformer Convolution.}
\label{vit-s}
\end{center}
\vskip -0.3in
\end{figure*}

\paragraph{ViT-small U-Net}\label{vit_unet_desc}
For the ViT-based U-Net, we use a ViT-small backbone. Our ViT-small has a patch size of 8, a latent dimension size of 384, an MLP head dimension of 1536, six attention heads, and twelve transformer blocks. Since U-Nets utilize downsampling feature maps during the upsampling, we pad our ViT-based U-Net with four downsampling blocks and four upsampling blocks. Maxpooling and double convolution follow each downsampling block. Similarly, each upsampling block contains bi-linear upsampling with double convolution. We present our ViT-small U-Net diagrammatically in Figure \ref{vit-s}. The ViT-Convolution (VC) block contains a convolution layer (kernel size =1 and no padding) and a linear layer. This convolution and linear layer convolution is preceded and followed up by a resize operation. The out convolution layer contains one convolution layer with no padding and kernel size = 1.  

For all our experiments we use a single Nvidia RTX-8000 GPU, equipped with 48 GB of video memory, 5 CPU cores and 64 GB system RAM. All experiments are repeated five times with different seed values unless state otherwise. 

\section{Additional Ablation Results}
\label{additional-abl-section}

\subsection{Change in loss function to cross-entropy}
\label{ce-ablation-expl}
As elucidated in Section \ref{section-loss-abl}, It is crucial to have a loss function that doesn't over-fit the majority/dominant class is important. We presented our rationale for using Focal-Tversky loss along with empirical results. Additionally, we also noted cross-entropy tends to overfit the dominant class. In this section we tested this empirically with MedSASS. We trained MedSASS end-to-end on the ISIC-2017 \cite{codella2017skin} dataset for 50 epochs and a batch size of 16. We only swapped the Focal-Tversky loss for a cross-entropy loss while keeping everything else constant. We obtained an IoU on the test set of 0 averaged over five seed values. To investigate further, we observed that the accuracy of the same model averaged over five seed values was 0.7608±0.002. Representing IoU and accuracy in terms of: TP: True Positive, TN: True Negative, FP: False Positive, and FN: False Negative, we can rewrite IoU and accuracy as: 

$IoU = \frac{TP}{TP+FN+FP}$ and $accuracy = \frac{TP+TN}{TP+FN+TN+FP}$. 

Since, IoU is 0 hence $TP$ is 0, and since accuracy is non-zero we know that $TP+TN$ has to be non-zero. Given the zero IoU, there are no true positives, meaning the model hasn't correctly identified and localized any of the objects of interest. Since, $TP$ is 0, $TN$ has to be non-zero. Hence, the architecture has overfit to the background in this case and is predicting background everywhere.

\subsection{Change in pre-training epoch and batch size}
\label{eph-bs-section}

In Table \ref{epcohs-abl}, our investigation focuses on the impact of varying the number of pre-training epochs on the performance of MedSASS, utilizing both ResNet-50 and ViT-small backbones in an end-to-end configuration. Additionally, Tables \ref{bs-abl-cnn} and \ref{bs-abl-vit} are dedicated to exploring how changes in batch size influence MedSASS when trained solely with encoder pre-training. We note that MedSASS demonstrates remarkable resilience against fluctuations in pre-training epochs. For change in the pre-training batch size we observe that although the performance remains constant mostly, there is a slight improvement when we reduce the batch size to 8. This attribute is particularly beneficial for practitioners operating under computational constraints, such as those in developing countries, enabling them to effectively deploy MedSASS. Furthermore, the robust performance of MedSASS with smaller datasets is advantageous for medical image analysis, especially in urgent scenarios like emerging pandemics or diseases lacking established data collection pipelines, such as autoimmune diseases. This capability of MedSASS can significantly contribute to enhancing existing diagnostic and treatment solutions.
\begin{table}[!hb]
\centering
\begin{tabular}{lllll}
\toprule
Backbone & \multicolumn{4}{c}{IoU on Test after pre-training (Epochs)}                                                                                                                                                                                         \\
             & \multicolumn{1}{c}{10}                                      & \multicolumn{1}{c}{20}                                      & \multicolumn{1}{c}{50}                                     & \multicolumn{1}{c}{70}                                     \\
             \midrule
ResNet50     & \begin{tabular}[c]{@{}l@{}}0.3785 \\ ±0.000249\end{tabular} & \begin{tabular}[c]{@{}l@{}}0.3778 \\ ±0.000568\end{tabular} & \begin{tabular}[c]{@{}l@{}}0.3824 \\ ±0.00579\end{tabular} & \begin{tabular}[c]{@{}l@{}}0.3876 \\ ±0.00098\end{tabular} \\
ViT-small        & \begin{tabular}[c]{@{}l@{}}0.3575 \\ ±0.04\end{tabular}     & \begin{tabular}[c]{@{}l@{}}0.3564 \\ ±0.0395\end{tabular}   & \begin{tabular}[c]{@{}l@{}}0.3633 \\ ±0.0062\end{tabular} & \begin{tabular}[c]{@{}l@{}}0.4091\\  ±0.0201\end{tabular} \\
\bottomrule
\end{tabular}
\caption{In this table, we present IoU averaged over five seed values for encoder-only training on the TissueNet dataset. We observe that the performance of CNNs is almost constant, even with the change in epoch. On the other hand, ViT-small performance increases with an increase in the number of pre-training epochs. This is in line with existing work involving Vision Transformers; this increase in performance is due to the nature of ViTs. While CNN have certain inductive priors, ViTs don't have them and learn during training.}
\label{epcohs-abl}
\end{table}

\begin{table}[h!]
     \begin{center}
     \begin{tabular}{  c  c  c }
     \toprule
      Input & Ground Truth & MedSASS Prediction \\ 
    \cmidrule(r){1-1}\cmidrule(lr){2-2}\cmidrule(l){3-3}%
     \raisebox{-\totalheight}{\includegraphics[width=0.23\textwidth]{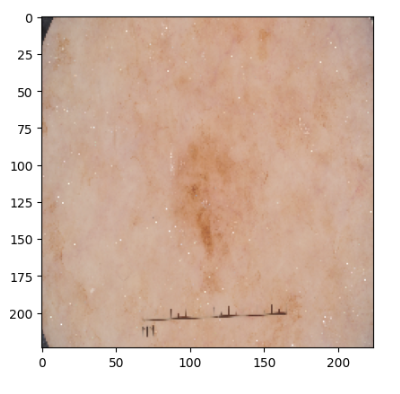}}
      & 
      \raisebox{-\totalheight}{\includegraphics[width=0.23\textwidth]{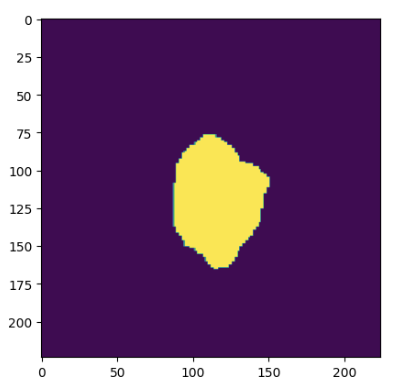}}
      &
      \raisebox{-\totalheight}{\includegraphics[width=0.25\textwidth]{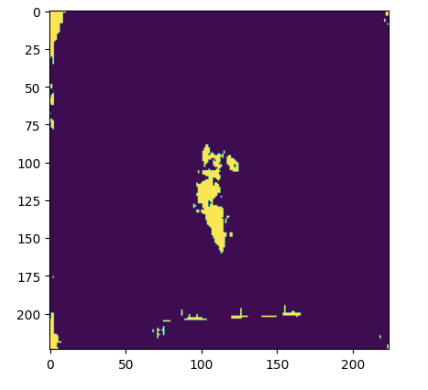}} \\ 
      \raisebox{-\totalheight}{\includegraphics[width=0.23\textwidth]{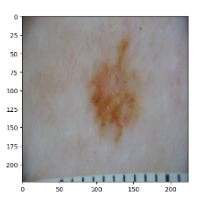}}
      & 
      \raisebox{-\totalheight}{\includegraphics[width=0.23\textwidth]{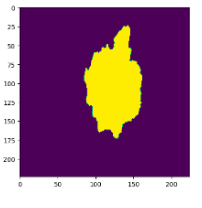}}
      &
      \raisebox{-\totalheight}{\includegraphics[width=0.23\textwidth]{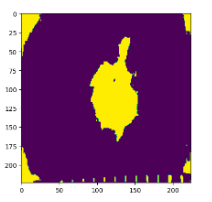}} \\ 
      \raisebox{-\totalheight}{\includegraphics[width=0.23\textwidth]{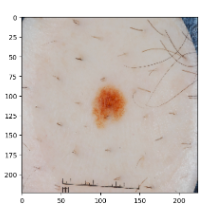}}
      & 
      \raisebox{-\totalheight}{\includegraphics[width=0.23\textwidth]{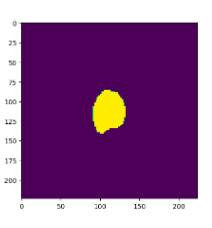}}
      &
      \raisebox{-\totalheight}{\includegraphics[width=0.23\textwidth]{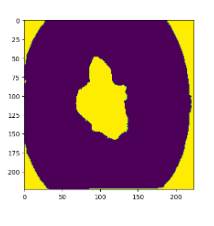}} \\ 
      \bottomrule
      \end{tabular}
      \caption{This table showcases three samples randomly chosen from the ISIC-2017 dataset test set, along with their corresponding ground truth masks and the predictions made by MedSASS CNN-based models. In the first row's sample, the MedSASS model underestimates the lesion area. Upon calculating the IoU for these predictions against the ground truth, MedSASS achieves a higher score of 0.5473 for the first sample. The second row demonstrates a closer prediction to the ground truth, with MedSASS models attaining IoU scores of 0.6790. For the final example, MedSASS acheives an IoU of 0.5085. Notably, all the outputs are from MedSASS trained only with self-supervision and no fine-tuning.}
      \label{tbl:myLboro}
      \end{center}
      \end{table}

\begin{table}[]
\centering
\begin{tabular}{l|lll}
\toprule
Dataset   & \multicolumn{3}{c}{Batch Size}                                                                                                                                                        \\
          & 8                                                           & 16                                                         & 32                                                         \\
          \midrule
ISIC-2017 & \begin{tabular}[c]{@{}l@{}}0.2551 \\ ±0.0243\end{tabular}   & \begin{tabular}[c]{@{}l@{}}0.2445 \\ ±0.00482\end{tabular} & \begin{tabular}[c]{@{}l@{}}0.2449 \\ ±0.00968\end{tabular} \\
\midrule
TissueNet       & \begin{tabular}[c]{@{}l@{}}0.3884\\  ±0.000241\end{tabular} & \begin{tabular}[c]{@{}l@{}}0.3824\\  ±0.00579\end{tabular} & \begin{tabular}[c]{@{}l@{}}0.3852\\  ±0.00607\end{tabular} \\
\bottomrule
\end{tabular}
\caption{This table presents the Intersection over Union (IoU) averaged across five distinct seed values for the encoder-only training of MedSASS employing a ResNet-50 architecture on the ISIC-2017 and TissueNet datasets, across varying batch sizes. The empirical evidence suggests remarkable stability in performance metrics, notwithstanding alterations in batch size.}
\label{bs-abl-cnn}
\end{table}

\begin{table}[]
\centering
\begin{tabular}{l|lll}
\toprule
Dataset   & \multicolumn{3}{c}{Batch Size}                                                                                                                                                    \\
          & 8                                                        & 16                                                         & 32                                                        \\
          \midrule
ISIC-2017 & \begin{tabular}[c]{@{}l@{}}0.2212 \\ ±0.131\end{tabular} & \begin{tabular}[c]{@{}l@{}}0.1914 \\ ±0.0839\end{tabular}  & \begin{tabular}[c]{@{}l@{}}0.2014 \\ ±0.0839\end{tabular} \\
\midrule
TissueNet       & \begin{tabular}[c]{@{}l@{}}0.413 \\ ±0.033\end{tabular}  & \begin{tabular}[c]{@{}l@{}}0.3633 \\ ±0.00624\end{tabular} & \begin{tabular}[c]{@{}l@{}}0.3822 \\ ±0.0204\end{tabular} \\
\bottomrule
\end{tabular}
\caption{In this table, we present the average IoU over five seed values for encoder-only training of MedSASS with ViT-small backbone on the ISIC-2017 and the TissueNet dataset with different batch sizes. As opposed to CNN, with ViT-small performance improves as we change the batch size, and best performance is observed with a batch size of 8.}
\label{bs-abl-vit}
\end{table}

\section{Comparison with Supervised approach}
\label{supervised-smp-app}
\begin{figure*}[!h]
\begin{center}
\centerline{\includegraphics[width=0.8\linewidth]{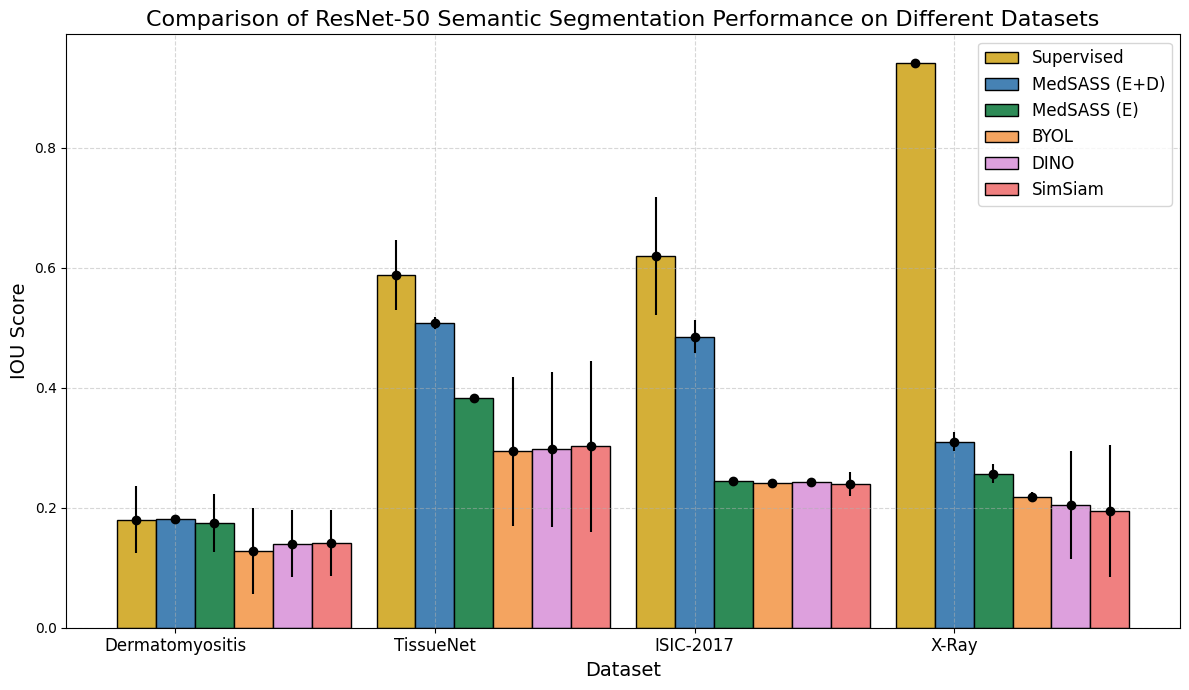}}
(a)
\centerline{\includegraphics[width=0.8\linewidth]{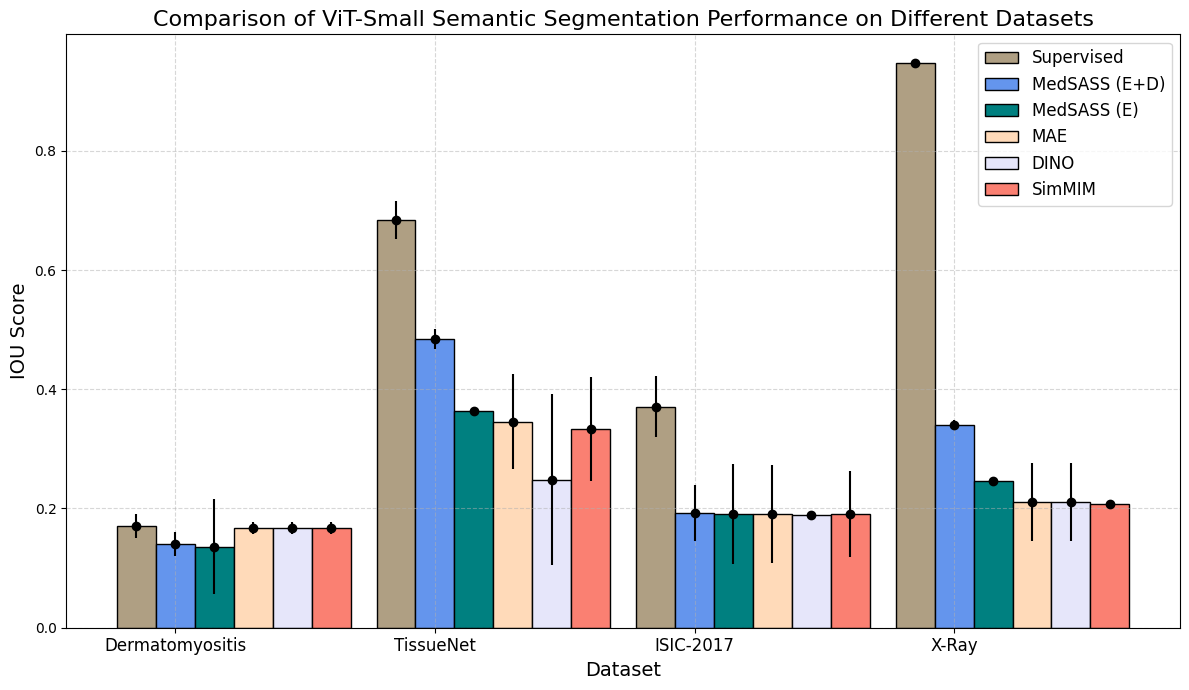}}
\centerline{(b)}
\caption{This figure provides a comparative analysis of different training methodologies: traditional supervised training, MedSASS with both encoder and decoder (MedSASS (E+D)), MedSASS with only the encoder (MedSASS (E)), and other leading self-supervised techniques. The black dot atop each bar, connected by a horizontal line, signifies the error bar derived from averaging over five different seed values. We present the result for CNN (ResNet-50) and ViT-small backbones in part (a) and (b) respectively of the figure. Furthermore, we note a significant performance disparity between supervised training and current state-of-the-art self-supervised techniques, with MedSASS generally bridging this performance gap.}
\label{all-everything}
\end{center}
\vskip -0.3in
\end{figure*}

In addition to comparing MedSASS to other self-supervised approaches for medical image segmentation, we also compared MedSASS with supervised learning. For a fair comparison, we train supervised approaches with focal-tversky loss. Additionally, all approaches were trained for 50 epochs with a batch size of 16. We present these results in Figure \ref{all-everything}. All self-supervised approaches are trained in a label-free manner, i.e., with 0\% labels in training, while the supervised approach was trained with 100\% labels.

We note that, aside from the Dermatomyositis dataset, there is a significant performance gap between supervised and self-supervised methods for segmentation. In most instances, MedSASS successfully bridges this gap to some extent, aligning the performance of self-supervised approaches with that of supervised methods.

\section{Image Classification Performance}
\label{classification-app}
As an additional comparison task, we also benchmark MedSASS and other state-of-the-art self-supervised approaches for classification tasks. For this, we use the ISIC-2017 and the X-ray datasets. For this task, we train the encoder for all the techniques, followed by linear probing for multi-class classification. We average the Recall score over the test set for five seed values and present the results in Table \ref{classification}.

\paragraph{Classification Task Details}

\begin{itemize}
    \item \textbf{X-Ray dataset} Classification task for the X-ray dataset is a multi-class classification task. Each image is classified into one of the four disease categories: normal, lung opacity, COVID, and viral pneumonia. We use the same training, validation, and testing splits as used for segmentation. 
    \item \textbf{ISIC-2017 dataset} Similar to the X-ray dataset task, the classification on the ISIC-2017 dataset is a multi-class classification out of three possible classes: melanoma, nevus-seborrheic keratosis, and healthy. We used the official training, validation, and testing splits of the dataset.
\end{itemize}

\paragraph{Experimental Details}


In this study, we employ self-supervised, pre-trained encoders from both CNN and Transformer architectures, followed by linear probing to assess their representational capabilities. Linear probing is conducted for 50 epochs, incorporating an early stopping criterion with a patience of five epochs to mitigate overfitting. To address the challenge of dataset imbalance, focal loss is strategically utilized. Evaluation is based on the recall score, an effective proxy for the class-balanced F1 score, to ensure equitable performance assessment across classes. For augmentation, use the following sequence: resize to (224,224) $\rightarrow$ (color jitters/random perspective) $\rightarrow$ (color jitter/random affine) $\rightarrow$ random vertical flip $\rightarrow$ random horizontal flip. All experiments were conducted on a single NVIDIA RTX8000 GPU with a batch size of 16.

\begin{table*}[]
\centering
\begin{tabular}{llll}
\hline
\textbf{Arch.}                      & \textbf{Approach} & \textbf{Recall on ISIC-2017} & \textbf{Recall on X-Ray} \\
\hline
\multirow{4}{*}{ResNet-50} & BYOL              & 0.65 ±0.00815                & \textbf{0.7073±0.0384}   \\
                           & DINO              & \underline{0.6546±0.00721}               & 0.5241±0.0701            \\
                           & SimSiam           & 0.6475 ±0.0104               & \underline{0.6541 ±0.0466}          \\
                           & MedSASS(e)        & \textbf{0.655 ±0.00845}      & 0.5785±0.0017            \\
                           \hline
\multirow{4}{*}{ViT-small}     & MAE               & 0.6492±0.0101                & \textbf{0.673±0.0173}    \\
                           & SimMIM            & 0.6438±0.0153                & \underline{0.66±0.0136}           \\
                           & DINO              & \underline{0.650±0.0117}                 & 0.40±0.0153              \\
                           & MedSASS(e)        & \textbf{0.6554±0.0087}       & 0.183±0.0255     \\
                           \hline
\end{tabular}
\caption{In this table, we present the linear probing results of encoder-only training of MedSASS and other state-of-the-art self-supervised approaches. We observe that for the ISIC-2017 dataset, both CNN and Transformer MedSASS outperform other self-supervised techniques, albeit marginally. On the X-ray dataset, it lags considerably behind in classification performance. The \textbf{bolded} numbers are best reported performance for a particular dataset and architecture, while the \underline{underlined} are the next best.}
\label{classification}
\end{table*}
\section{Drone dataset Performance}
\label{natural-datasets}

The drone dataset \cite{semantic_drone_dataset} features urban landscapes captured from a nadir perspective, showcasing over 20 residential structures at heights ranging from 5 to 30 meters. High-resolution imagery was obtained using a camera that produces 6000x4000 pixel images. This dataset is richly annotated, containing objects categorized into 23 semantic classes such as unlabeled, paved-area, dirt, grass, gravel, water, rocks, pool, vegetation, roof, wall, window, door, fence, fence-pole, person, dog, car, bicycle, tree, bald-tree, ar-marker, obstacle, and conflicting. We undertake a semantic segmentation task where we differentiate relevant areas—signified by the presence of any objects from the 23 categories—from the background. The outcomes of these experiments are detailed in Sections \ref{binary-drone}.

We train encoder-only ResNet-50 methods for 50 epochs with a batch size of 16 and an early-stopping patience of 5 epochs for both binary and multi-class semantic segmentation. We averaged the mean IoU, over three seed values on the test set.

\subsection{Semantic segmentation}
\label{binary-drone}
We present the results of encoder-only training on the drone dataset in Table \ref{binary-drone-results}. In this table, we compare the performance of MedSASS and other state-of-the-art self-supervised techniques over three seed values. We used ResNet-50 as our choice of encoder and pre-trained all approaches for 50 epochs.
\begin{table}[]
\centering
\begin{tabular}{lll}
\hline
Approach & Arch. & IoU                    \\
\hline
SimSiam  & ResNet-50 & \textbf{0.9932±0.0345} \\
BYOL     & ResNet-50 & 0.9910±0.0613          \\
MedSASS  & ResNet-50 & 0.9744±0.0445          \\
DINO     & ResNet-50 & 0.9459±0.0527         \\
\hline
\end{tabular}
\caption{In this table, we present the IoU averaged over three seed values on the test set of the drone dataset. In this task, we performed binary semantic segmentation, wherein we had to segment the objects of interest/foreground (paved area, dirt, grass, gravel, water, rocks, pool, vegetation, roof, wall, window, door, fence, fence-pole, person, dog, car, bicycle, tree, bald-tree, ar-marker, obstacle, and conflicting) from the background. As natural images do not have the same priors as medical images, extracting labels for supervision using Otsu's approach becomes difficult. Hence, MedSASS is unable to keep up with the state-of-the-art self-supervised technique in this case, but at the same time, it outperformed DINO.}
\label{binary-drone-results}
\end{table}

\section{On the efficacy of Otsu's approach}
\label{samples}
Since MedSASS relies on Otsu's approach for self-supervision, the quality of output from Otsu's approach is crucial. In this section, we provide samples of input images, their corresponding ground truths, and their respective output from Otsu's approach. 

\subsection{Dermatomyositis Dataset}
\begin{figure}[!h]
\begin{center}
\centerline{\includegraphics[width=0.8\linewidth]{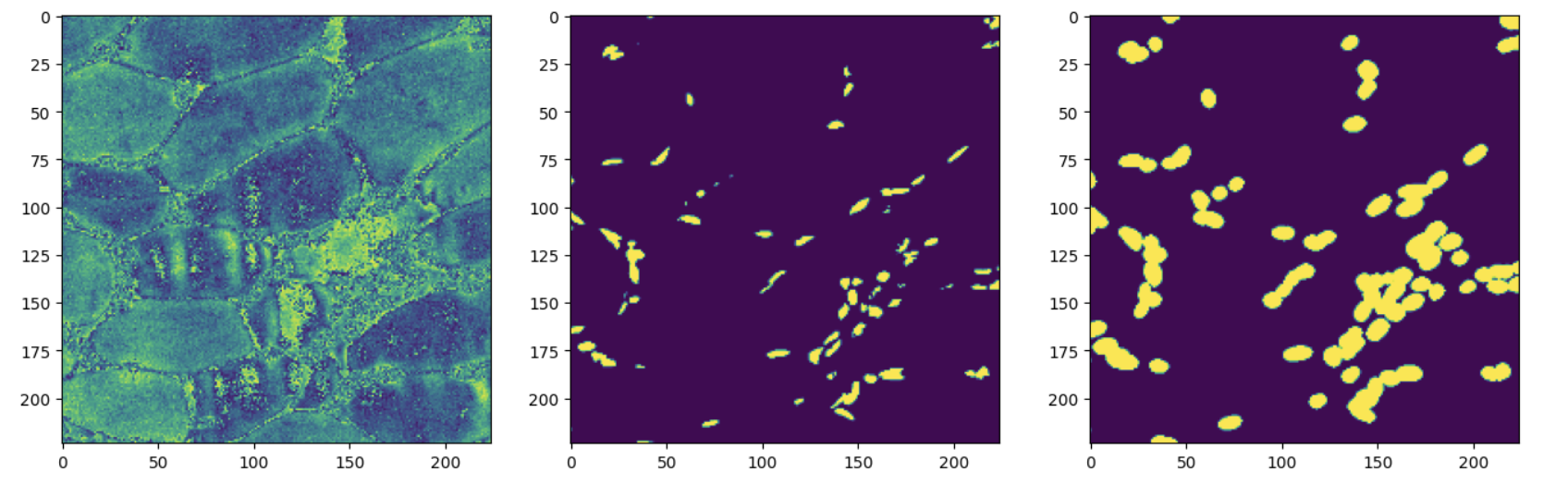}}
\centerline{\includegraphics[width=0.8\linewidth]{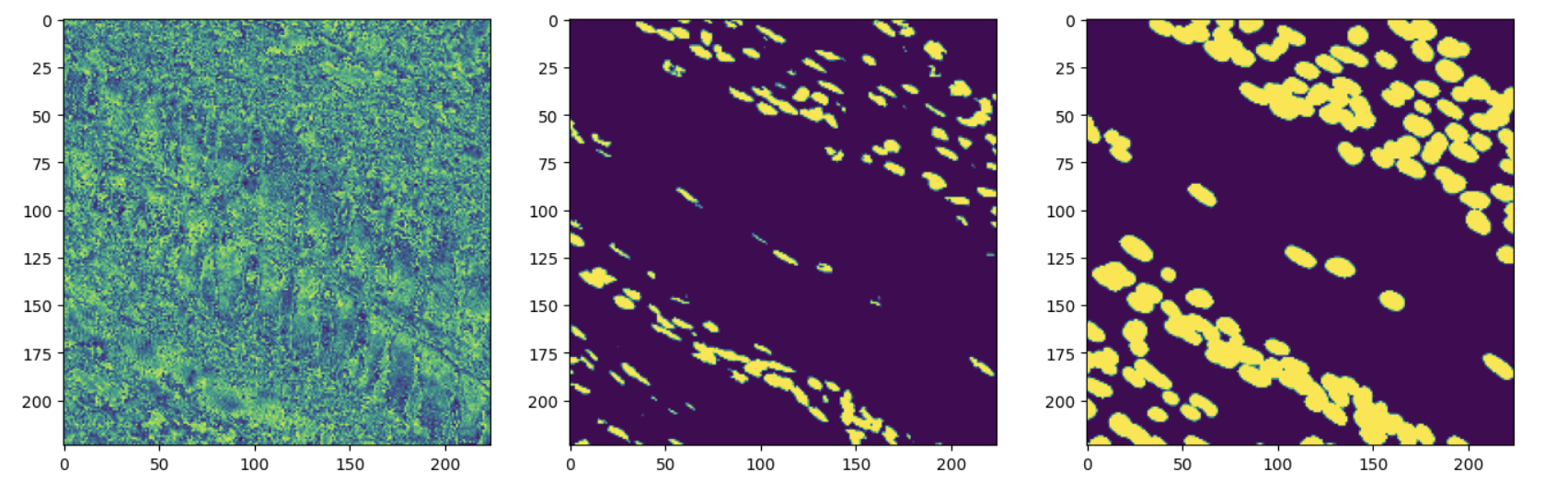}}
\caption{In this figure, we show two samples from the Dermatomyositis dataset \cite{singh2023data, VANBUREN2022113233}. In the rightmost column, we have the input image; in center column, we present the output from Otsu's approach, and finally, the corresponding ground truth image is in the rightmost column.}
\label{derm-Otsu-output}
\end{center}
\vskip -0.3in
\end{figure}

We present images, the corresponding labels, and the output from Otsu's approach for samples from the Dermatomyositis dataset in Figure \ref{derm-Otsu-output}. We observe that while the output from Otsu's approach is not identical to the ground truth images, they are similar enough to be used for supervision instead of the actual ground truth images. Additionally, the Dermatomyositis dataset is a challenging dataset due to the complexity and large number of small objects present in the dataset (as presented in Fig. \ref{derm-Otsu-output}), along with the limited number of samples available in the dataset.

\subsection{ISIC-2017 Dataset}
\begin{figure}[!h]
\begin{center}
\centerline{\includegraphics[width=0.8\linewidth]{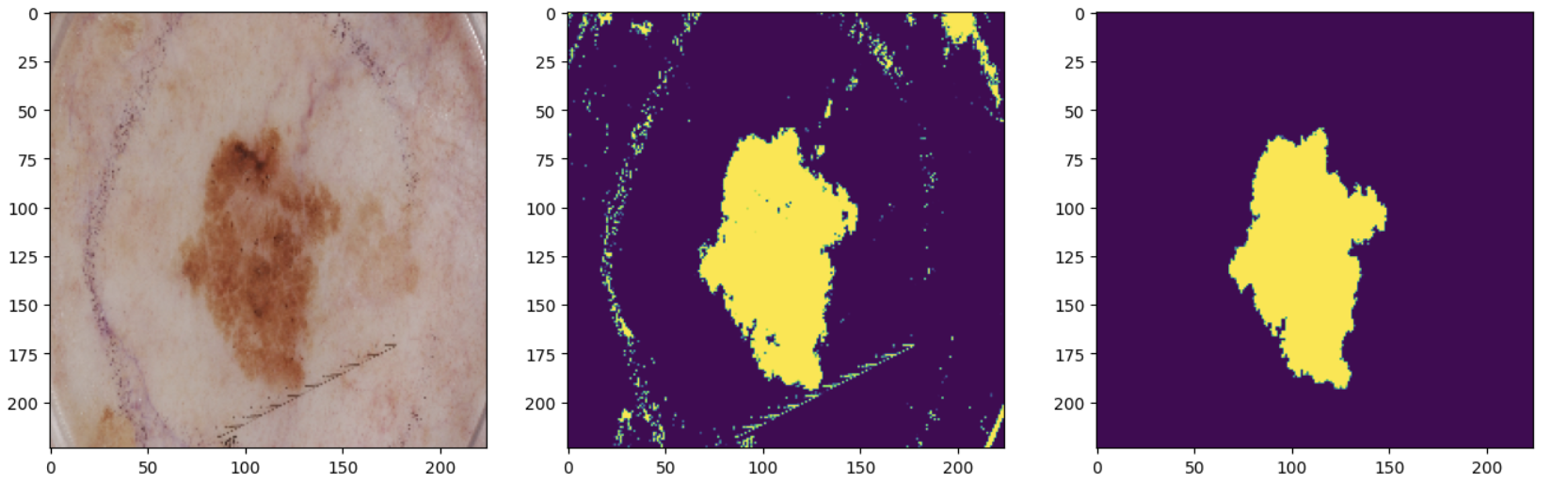}}
\centerline{\includegraphics[width=0.77\linewidth]{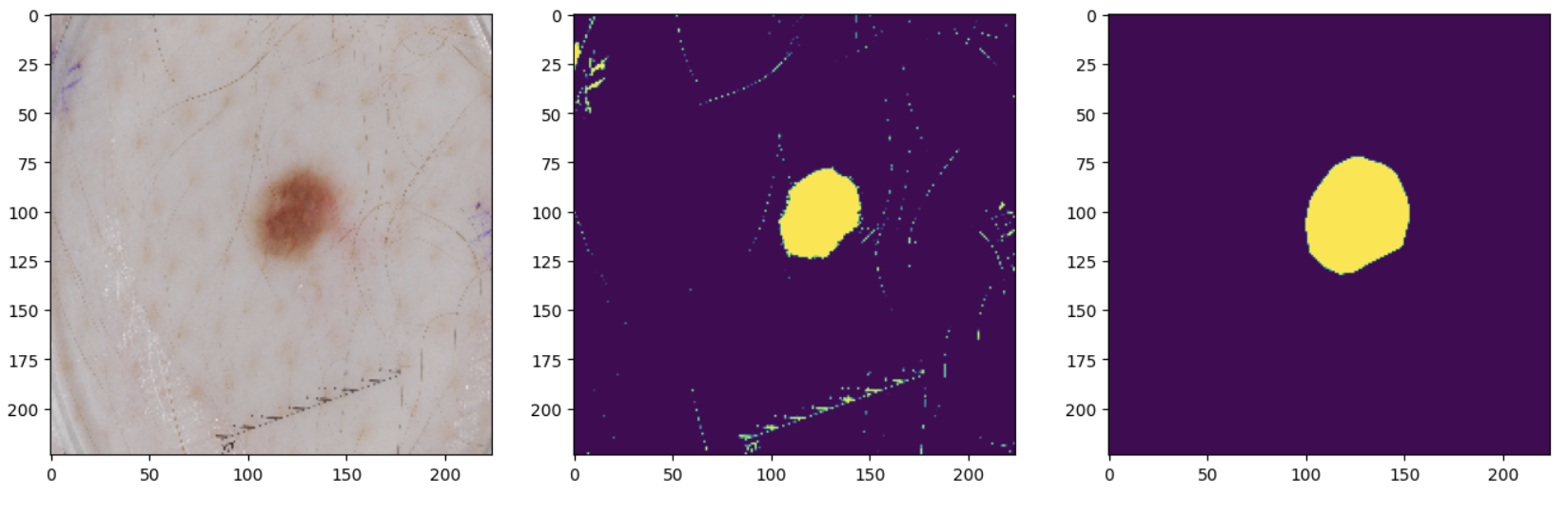}}
\caption{In this figure, we show two samples from the ISIC-2017 \cite{codella2017skin} dataset. In the rightmost column, we have the input image; in center column, we present the output from Otsu's approach, and finally, the corresponding ground truth image is in the rightmost column.}
\label{isic-Otsu-output}
\end{center}
\vskip -0.3in
\end{figure}

Similarly, we present samples from the ISIC-2017 dataset in Figure \ref{isic-Otsu-output}. The ISIC-2017 dataset is a particularly challenging dataset due to small inter-class variation and obfuscation of the lesion region \cite{singh2023data}. 

\subsection{Tissuenet Dataset}

\begin{figure}[!h]
\begin{center}
\centerline{\includegraphics[width=0.8\linewidth]{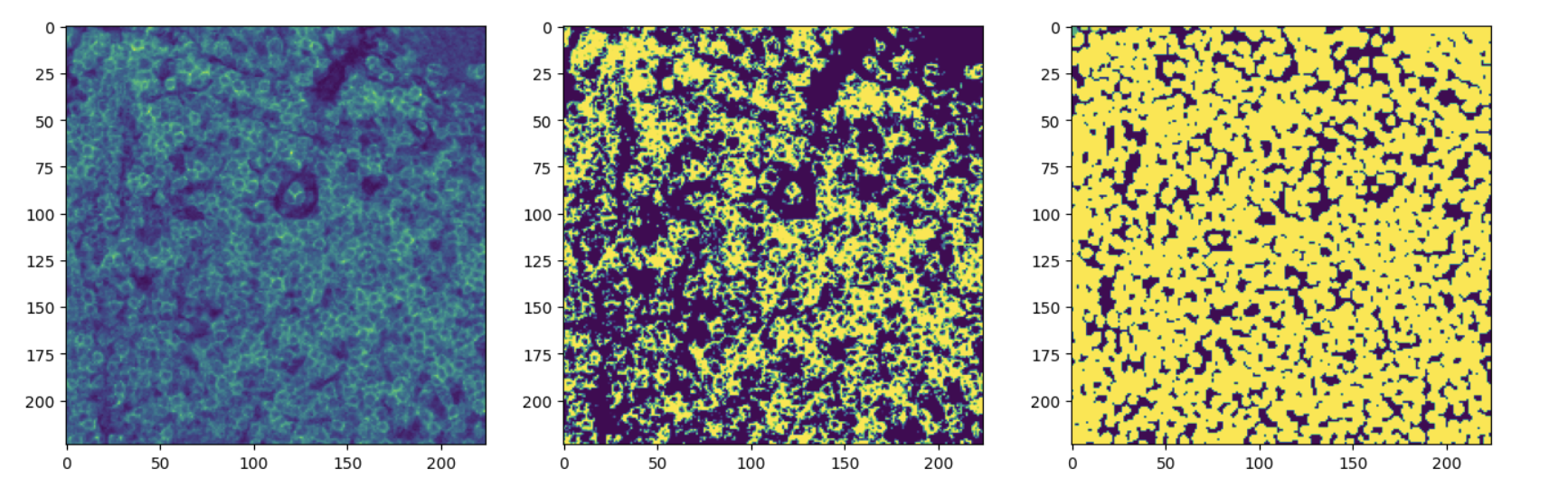}}
\centerline{\includegraphics[width=0.77\linewidth]{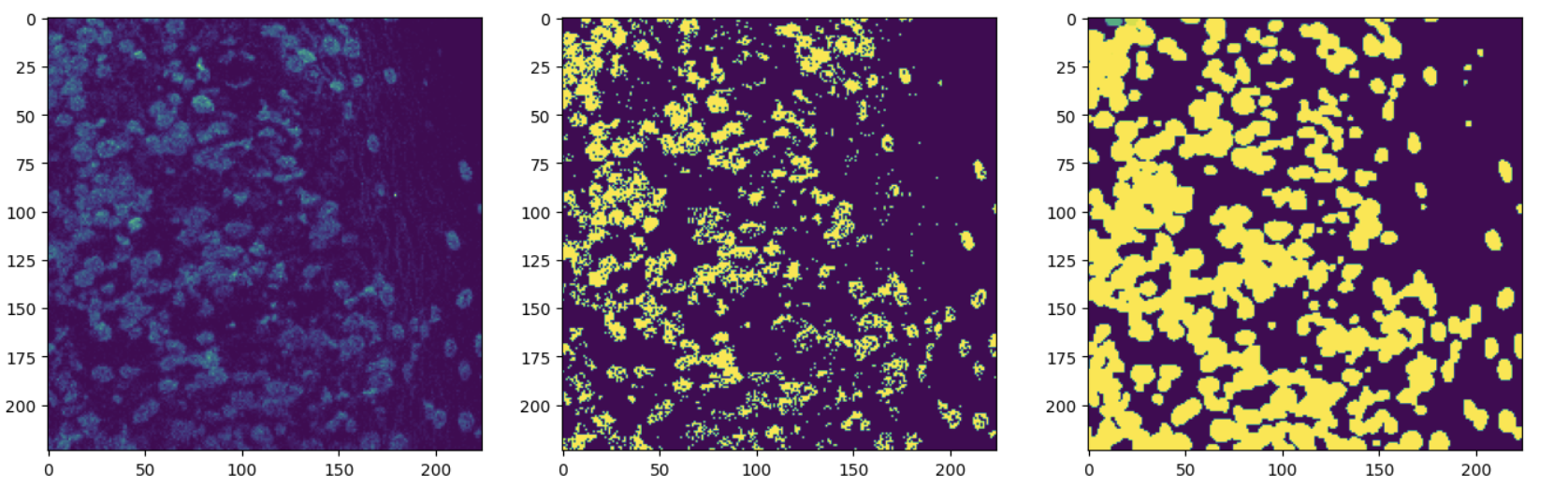}}
\caption{Sample images from the TissueNet dataset \cite{greenwald2022whole}, similar to Fig. \ref{derm-Otsu-output} and \ref{isic-Otsu-output}, in the leftmost column we have the input images, output from Otsu in middle and ground truth corresponding to the input image in the rightmost column.}
\label{tn-Otsu-output}
\end{center}
\vskip -0.3in
\end{figure}

In Fig. \ref{tn-Otsu-output}, we present the input image in the leftmost column, Otsu's output in the middle, and the corresponding ground truth for the input image in the rightmost column. Similar to the Dermatomyositis dataset, this dataset also contains a large number of small objects. The TissueNet dataset came from a number of different imaging platforms, such as CODEX, Cyclic Immunofluorescence, IMC, MIBI, Vectra, and MxIF. It shows a wide range of disease states and tissue types. Furthermore, this dataset consists of paired nuclear and whole-cell annotations, which, when combined, sum up to more than one million paired annotations. \cite{greenwald2022whole}.

\subsection{X-Ray Dataset}

\begin{figure*}[!h]
\begin{center}
\centerline{\includegraphics[width=0.8\linewidth]{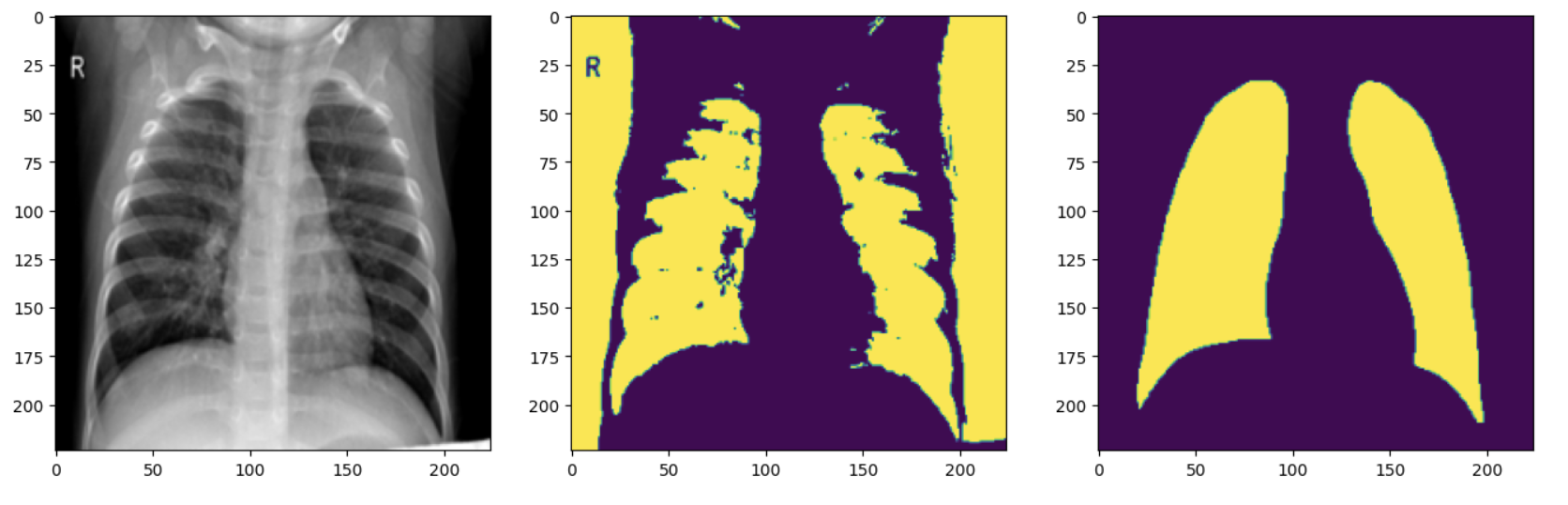}}
\centerline{\includegraphics[width=0.8\linewidth]{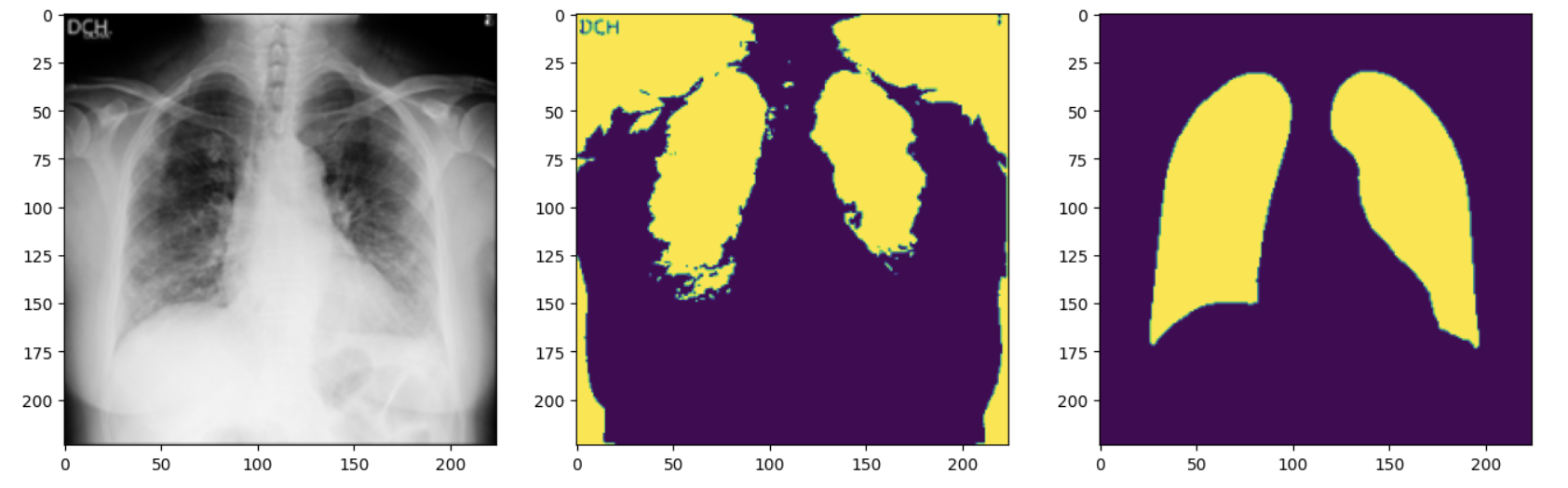}}
\caption{Sample images from the X-ray dataset \cite{chowdhury2020can,rahman2021exploring} are displayed as follows: the leftmost column features the input images, the center column presents the output from Otsu, and the rightmost column showcases the ground truth corresponding to each input image.}
\label{xray-Otsu-output}
\end{center}
\vskip -0.3in
\end{figure*}

X-ray imaging stands as a predominant diagnostic tool due to its widespread availability, cost-effectiveness, non-invasiveness, and simplicity. But lung segmentation is hard because of three things: (1) changes that are not the result of disease, such as changes in lung shape and size due to age, gender, and heart size; (2) changes that are the result of disease, such as high-intensity opacities brought on by severe pulmonary conditions; and (3) obfuscation in imaging, such as clothing or medical devices (pacemakers, infusion tubes, catheters) that cover the lung field. We present two samples from the X-ray dataset in Fig. \ref{xray-Otsu-output} along with their Otsu's output and corresponding ground truth. Despite the challenges for each dataset we observe that Otsu's labels provide non-blank and masks that are similar to the ground truth. 

\section{Description of Metrics}

\paragraph{For segmentation}

For measuring segmentation performance, we use the IoU, or Intersection Over Union, metric. It helps us understand how similar the sample sets are.\\ 
\begin{center}
$IoU = \frac{\text{area of overlap}}{\text{area of union}}$
\end{center}

\paragraph{For classification} we used the recall score as our comparison metric for multi-class classification, which is defined as $Recall = \frac{TP}{TP+FN}$, where TP: True Positive, TN: True Negative, FP: False Positive, and FN: False Negative. 
Since, both the datasets have class imbalance to get a real sense of performance we use Recall value. This is because it is semantically equal to the balanced multi-class accuracy value,
 TP: True Positive, TN: True Negative, FP: False Positive, and FN: False Negative.

\end{document}